\documentclass{article} 

\usepackage{iclr2027_conference,times}

\usepackage{amsmath,amsfonts,bm}

\def\eqref#1{equation~\ref{#1}}

\def\1{\bm{1}}

\DeclareMathAlphabet{\mathsfit}{\encodingdefault}{\sfdefault}{m}{sl}
\SetMathAlphabet{\mathsfit}{bold}{\encodingdefault}{\sfdefault}{bx}{n}

\usepackage{amsmath}
\usepackage{amssymb}
\usepackage{amsthm}

\usepackage{algorithm}
\usepackage{algpseudocode}

\usepackage{hyperref}
\usepackage{url}

\newtheorem{theorem}{Theorem}
\newtheorem{proposition}{Proposition}

\newtheorem{assumption}{Assumption}

\usepackage{float}
\usepackage{caption}
\usepackage{graphicx}
\usepackage{array}
\usepackage{booktabs}
\usepackage{multirow}
\usepackage{microtype}

\usepackage{placeins}

\newcolumntype{C}[1]{>{\centering\arraybackslash}m{#1}}


\title{Constrained Variational Policy Optimization for Hybrid Action Spaces}

\author{Shuchang Yan \\
\texttt{ycyan@connect.hku.hk}}

\iclrfinalcopy

\begin{document}

\maketitle
\lhead{Preprint}

\begin{abstract}
Constrained reinforcement learning in hybrid action spaces must coordinate
discrete modes and mode-conditioned continuous controls under long-horizon
cost constraints. Existing constrained hybrid-action methods typically
optimize factorized policies directly in parameter space, whereas
Constrained Variational Policy
Optimization (CVPO) separates nonparametric improvement from parametric
projection in its original continuous-action formulation.
We introduce Hybrid
Constrained Variational Policy Optimization (H-CVPO), extending this
variational framework to hybrid action spaces. Under boundedness and strict
feasibility, its constrained E-step has zero duality gap and, when the dual
optimum is attained at positive temperature, yields a closed-form joint target
distribution over hybrid actions through exponential reward--cost
reweighting. Using the exact hybrid KL decomposition, we formulate a
factorized M-step that fits both policy components to this target
under separate discrete and conditional-continuous trust regions. For exact
E/M updates, we prove fixed-reference surrogate improvement and conditional
reward--cost bounds with a sufficient feasibility-margin condition. For
approximate updates, we establish the existence of a dual minimizer under
continuity on a compact domain and use residual-based admissibility to
determine whether to project the target. Ablations identify cost-aware
reweighting as an important contributor to constraint control. Extensive
experiments in simulated hybrid-control environments show that H-CVPO
achieves favorable reward--cost trade-offs compared with hybrid-action baselines.
\end{abstract}

\section{Introduction}

Many real-world control systems are inherently hybrid: each decision
couples a discrete operating mode with a mode-conditioned continuous control
action. Hybrid electric vehicles illustrate this structure: a multi-mode
powertrain selects an electric or engine-assisted mode and continuously
allocates power between the engine and electric machines
~\citep{Zhuang2017ModeShift,Fan2021ModeSelection}. Similar structures arise in
power systems, where unit commitment determines generator on/off states and
economic dispatch sets continuous generation levels
~\citep{Montero2022ReviewUC,Swarup2006UCED}; in steel-plant oxygen supply, where
production scheduling combines equipment start-up and shutdown decisions
with continuous oxygen output
~\citep{Li2023ManufacturingHybrid}; in autonomous driving, where decisions to
keep or change lanes are coupled with continuous longitudinal acceleration
~\citep{Lin2024LaneChangeHybrid}; and in robotic assembly, where selecting
between coarse and fine motion control can be coupled with continuous
velocity commands
~\citep{Neunert2020HybridRobotics}. Such problems are neither purely discrete
nor purely continuous and require coordinated optimization of mode selection
and mode-conditioned control under operational constraints.

Constrained reinforcement learning (CRL) is commonly formalized as a
constrained Markov decision process (CMDP), which maximizes long-horizon
performance subject to cumulative cost
limits~\citep{altman1999cmdp}. Representative methods enforce these constraints
through primal--dual optimization~\citep{Tessler2018RCPO},
trust-region constrained updates~\citep{Achiam2017CPO}, projection-based
methods~\citep{Yang2020PCPO}, or Lyapunov-based approaches
~\citep{Chow2018Lyapunov}. These methods address constrained policy
optimization but do not explicitly account for factorized, mode-conditioned
hybrid actions or coordinate their discrete and continuous updates.
A complementary perspective on policy optimization comes from reinforcement
learning as probabilistic inference~\citep{levine2018rlasinference}. Maximum a Posteriori
Policy Optimization (MPO) constructs an improved nonparametric action
distribution and projects it onto a parametric policy through an
expectation--maximization (EM) procedure~\citep{Abdolmaleki2018MPO}.
Constrained Variational Policy Optimization (CVPO) incorporates constraints
through a nonparametric E-step and a KL-controlled M-step, connecting
constrained policy optimization and probabilistic inference~\citep{Liu2022CVPO}.

Extending this variational framework to hybrid control, however, requires
a joint nonparametric target over discrete modes and mode-conditioned
continuous controls. Projecting this target onto a factorized policy
requires divergence controls that account for the dependence between
mode selection and mode-conditioned continuous control.
Complementary to variational methods, CHPO addresses constrained hybrid
control by formulating constrained parameterized-action Markov decision
processes and optimizing factorized hybrid policies through
feasibility-dependent objective switching~\citep{Zhou2025CHPO}. Its updates
remain in parametric policy space rather than constructing a joint
nonparametric target, and do not provide separate divergence controls for
discrete-mode and continuous-action updates.

Building on the constrained variational framework, we develop Hybrid
Constrained Variational Policy Optimization (H-CVPO) for hybrid action
spaces. Its constrained E-step constructs a joint nonparametric target,
and its M-step projects this target onto a factorized policy under
separate discrete and conditional-continuous KL constraints.

Our contributions are summarized as follows:

\begin{itemize}

\item We formulate constrained variational policy improvement for hybrid
action spaces, extending the continuous-action formulation to a
joint nonparametric target that couples discrete mode selection with
mode-conditioned continuous control.

\item Under stated assumptions, we derive the resulting hybrid E/M update.
The constrained E-step has strong duality and, for attained positive-temperature
dual optima, yields a closed-form reward--cost-reweighted target. The exact
hybrid KL decomposition then yields a factorized M-step with separate
discrete and conditional-continuous trust regions.

\item We establish fixed-reference surrogate improvement for exact E/M
updates and conditional reward--cost bounds with a sufficient feasibility
margin. For approximate updates, we prove existence of a dual minimizer
under continuity on a compact domain and introduce residual-based
target-admissibility conditions.

\item We compare H-CVPO with hybrid-action baselines across four tasks
and three cost limits under matched interaction budgets, and use
component-wise ablations to examine the effects of cost-aware reweighting,
separate hybrid KL controls, and target-admissibility filtering
on reward and cost.

\end{itemize}

\section{Related Work}

H-CVPO builds on variational policy optimization to address constrained
reinforcement learning in hybrid action spaces.

\paragraph{Hybrid-action reinforcement learning.}

Hybrid-action methods couple a discrete choice with associated continuous
parameters. Q-PAMDP learns action-selection and parameter-selection
policies~\citep{masson2016parameterized}; PADDPG embeds parameterized actions in
deterministic actor--critic learning~\citep{Hausknecht2015PADDPG}; P-DQN combines
value-based discrete selection with continuous parameter optimization
~\citep{xiong2018pdqn}; and HPPO separates discrete and continuous policy
branches~\citep{Fan2019HPPO}. Hybrid SAC further extends entropy-regularized
actor--critic learning to hybrid actions~\citep{delalleau2019hybridsac}.
Neunert et al. extend MPO to native hybrid action spaces, jointly optimizing
discrete and continuous actions, and demonstrate the approach in robotic
control \citep{Neunert2020HybridRobotics}. These methods do not explicitly
impose cumulative cost constraints.

\paragraph{Constrained reinforcement learning.}

Within the CMDP framework, RCPO and PID Lagrangian methods handle constraints
through multiplier updates
~\citep{Tessler2018RCPO,stooke2020pidlagrangian}, whereas CPO and PCPO use
trust-region and projection-based updates
~\citep{Achiam2017CPO,Yang2020PCPO}. Their formulations do not explicitly
address the coupling between discrete mode selection and mode-conditioned
continuous control.

\paragraph{Variational policy optimization.}

Probabilistic-inference formulations derive policy-improvement updates from
inference objectives~\citep{levine2018rlasinference}. MPO alternates
nonparametric action-distribution improvement with parametric policy projection
~\citep{Abdolmaleki2018MPO}. CVPO incorporates cumulative constraints into this
E/M procedure through a constraint-aware E-step and a KL-controlled M-step
~\citep{Liu2022CVPO}. More recently, SafeMPO extends MPO-style nonparametric
policy improvement to CMDPs through a probabilistic incremental
safety-improvement constraint, followed by policy cloning
\citep{Mattick2026SafeMPO}. Its reported experiments use continuous action
spaces and do not explicitly represent discrete modes or mode-conditioned
continuous controls. Neither CVPO nor SafeMPO explicitly formulates a joint
hybrid-action target or its projection onto a factorized policy with separate
discrete and conditional-continuous KL constraints.

\paragraph{Constrained hybrid-action reinforcement learning.}

Under the assumptions of its analysis, CHPO derives finite-iteration bounds
on the policy-objective gap and the excess of expected cost over its
threshold~\citep{Zhou2025CHPO}. To complement this theoretical result, we
include two constrained hybrid baselines: HPPO-Lag and PDQN-Rco, formed by
augmenting HPPO and P-DQN with Lagrangian and reward-constrained updates,
respectively. We report their empirical reward and cost under the same
interaction budget used for H-CVPO.

\section{Preliminaries}
\label{sec:preliminaries}

\subsection{CMDP Formulation}

We consider a discounted CMDP~\citep{altman1999cmdp},
$\mathfrak{M}=(\mathcal{S},\mathcal{A},P,r,c,\rho_0,\gamma)$, with cumulative
cost limit $d$. Here, $\mathcal S$ and $\mathcal A$ are standard Borel state
and action spaces, $P$ is a stochastic transition kernel from
$\mathcal S\times\mathcal A$ to $\mathcal S$, $\rho_0$ is the initial-state
distribution, and $\gamma\in(0,1)$ is the discount factor.
The measurable reward and cost functions
$r:\mathcal S\times\mathcal A\rightarrow\mathbb R$ and
$c:\mathcal S\times\mathcal A\rightarrow\mathbb R_+$ satisfy
$|r(s,a)|\le R_{\max}$ and $0\le c(s,a)\le C_{\max}$ for finite
$R_{\max},C_{\max}$. We restrict attention to stationary randomized Markov
policies $\pi$, each a stochastic kernel from $\mathcal S$ to $\mathcal A$.

Under $\pi$, a trajectory
$\tau=(s_0,a_0,s_1,a_1,\ldots)$ is generated by
$s_0\sim\rho_0$, $a_t\sim\pi(\cdot\mid s_t)$, and
$s_{t+1}\sim P(\cdot\mid s_t,a_t)$. The expected discounted reward and cost returns are
\begin{equation}
J_r(\pi)
=
\mathbb{E}_{\tau\sim(\rho_0,P,\pi)}
\!\left[
\sum_{t=0}^{\infty}\gamma^t r(s_t,a_t)
\right],
\qquad
J_c(\pi)
=
\mathbb{E}_{\tau\sim(\rho_0,P,\pi)}
\!\left[
\sum_{t=0}^{\infty}\gamma^t c(s_t,a_t)
\right].
\end{equation}
The constrained policy-optimization problem is
\begin{equation}
\label{eq:cmdp_pre}
\max_{\pi}\;J_r(\pi)
\qquad
\mathrm{s.t.}
\qquad
J_c(\pi)\le d .
\end{equation}
A policy is feasible if $J_c(\pi)\le d$. We next specialize $\mathcal A$
to a hybrid action space.

\subsection{Hybrid Action Space and Policy Representation}

For hybrid control, the generic CMDP action space is instantiated as
$\mathcal{A}=\mathcal{H}$. A hybrid action $h=(m,a_c)$ consists of a discrete
mode $m$ and a mode-conditioned continuous action $a_c$. Let
$\mathcal{M}=\{1,\ldots,M\}$ be the finite mode set and let
$\mathcal{A}_c^m\subseteq\mathbb{R}^{p_m}$ be the Borel continuous action domain
for mode $m$, with mode-dependent dimension $p_m\in\mathbb N_0$ and positive
Lebesgue measure when $p_m>0$.
For $p_m=0$, we set $\mathcal A_c^m=\mathbb R^0$, a singleton, and
interpret $da_c$ as its unit mass, thereby covering modes without a free
continuous parameter. The hybrid action space is the disjoint union
\begin{equation}
\label{eq:hybrid_action_measure}
\mathcal{H}
=
\bigsqcup_{m\in\mathcal{M}}
\left(\{m\}\times\mathcal{A}_c^m\right).
\end{equation}
For any nonnegative measurable function
$\varphi:\mathcal{H}\rightarrow[0,\infty]$, define the sum--integral
reference measure $\mu$ by
\begin{equation}
\label{eq:hybrid_reference_measure}
\int_{\mathcal{H}}\varphi(h)\,\mu(dh)
=
\sum_{m\in\mathcal{M}}
\int_{\mathcal{A}_c^m}\varphi(m,a_c)\,da_c .
\end{equation}
Thus, $\mu$ combines summation over modes with Lebesgue integration on each
continuous action space, allowing its dimension and domain to depend on the
mode. We restrict the hybrid action distributions considered below to those
that are absolutely continuous with respect to $\mu$. Their densities, and
hence all KL divergences, are defined relative to $\mu$.

Accordingly, a stochastic hybrid policy is represented by its $\mu$-density
\begin{equation}
\label{eq:hybrid_policy_pre}
\pi_\theta(m,a_c\mid s)
=
\pi_{\theta_m}(m\mid s)\pi_{\theta_c}(a_c\mid s,m),
\end{equation}
where $\theta=(\theta_m,\theta_c)$ collects the discrete- and
continuous-policy parameters. For any measurable function
$f:\mathcal S\times\mathcal H\rightarrow\mathbb R$ that is integrable under
$\pi_\theta(\cdot\mid s)$, its conditional expectation is
\begin{equation}
\label{eq:hybrid_expectation_pre}
\begin{aligned}
\mathbb{E}_{(m,a_c)\sim\pi_\theta(\cdot\mid s)}[f(s,m,a_c)]
&=
\sum_{m\in\mathcal{M}}
\pi_{\theta_m}(m\mid s)
\int_{\mathcal{A}_c^m}
f(s,m,a_c) \\
&\qquad\qquad
\pi_{\theta_c}(a_c\mid s,m)\,da_c .
\end{aligned}
\end{equation}
Although factorized, the continuous component remains conditioned on the
selected mode, so the policy defines a joint distribution over hybrid actions.

When $\mathcal{A}_c^m=\mathcal{A}_c$ for every $m$, the hybrid action space
reduces to $\mathcal{M}\times\mathcal{A}_c$. If
$\pi_{\theta_c}(a_c\mid s,m)=\pi_{\theta_c}(a_c\mid s)$ as well, the policy
further reduces to an independent discrete--continuous factorization. Thus,
the formulation covers the general variable-domain case, the fixed-domain
special case, and the further special case of independent discrete--continuous
policies.

\subsection{Variational Policy Improvement}

Following MPO and CVPO~\citep{Abdolmaleki2018MPO,Liu2022CVPO}, variational
policy optimization introduces an intermediate nonparametric action
distribution between consecutive parametric policies. At iteration $i$, the
E-step optimizes a target $q_i^\star(\cdot\mid s)$ while holding the current
policy $\pi_i$
fixed, and the M-step projects this target onto the policy class to obtain
$\pi_{i+1}$. The resulting update is
\[
\pi_i \longrightarrow q_i^\star \longrightarrow \pi_{i+1}.
\]
This separation decouples target optimization from parametric policy fitting.
For hybrid actions, however, $q_i^\star(\cdot\mid s)$ is a probability measure
on $\mathcal H$
rather than a pair of independently optimized action distributions, so its
construction and projection must preserve the mode-conditioned policy
structure and the associated hybrid KL geometry.

\section{Hybrid Constrained Variational Policy Optimization}
\label{sec:hcvpo}

We now formulate the preceding variational update for the hybrid CMDP. The
constrained E-step constructs a joint nonparametric target
$q_i^\star(m,a_c\mid s)$ on $\mathcal H$ relative to the current policy
$\pi_i$, and the factorized M-step projects the same target onto a
parameterized hybrid policy.

\subsection{Hybrid-Action Variational E-Step}
\label{sec:mixed_estep}

At iteration $i$, fix a reference probability measure $d_i$ on $\mathcal S$,
the current hybrid policy $\pi_i$, and measurable reward and cost action-value
functions $Q_r^i,Q_c^i:\mathcal S\times\mathcal H\rightarrow\mathbb R$.
We write $\pi_i(m,a_c\mid s)$ for the $\mu$-density of
$\pi_i(\cdot\mid s)$. The reference measure $d_i$, current policy $\pi_i$,
and action-value functions $Q_r^i$ and $Q_c^i$ remain fixed throughout one E/M
cycle. Let $q$ be a measurable stochastic kernel from $\mathcal S$ to
$\mathcal H$ such that $q(\cdot\mid s)\ll\pi_i(\cdot\mid s)$ for $d_i$-almost
every (a.e.) state $s$, where $\ll$ denotes absolute continuity. The conditional KL
divergence is
$D_{\mathrm{KL}}(q\|\pi_i)=\int_{\mathcal H}\log(dq/d\pi_i)\,dq$, with the
state conditioning suppressed; since $\pi_i(\cdot\mid s)\ll\mu$, every
admissible $q(\cdot\mid s)$ also admits a $\mu$-density.
For an E-step cost threshold $\epsilon_{E,i}>0$ and KL radius
$\epsilon_{\mathrm{KL}}>0$, the E-step solves
\begin{equation}
\label{eq:hcvpo_estep}
\begin{aligned}
\max_q\quad&
\mathbb{E}_{s\sim d_i,\,h\sim q(\cdot\mid s)}
\left[Q_r^i(s,h)\right]
\\
\mathrm{s.t.}\quad&
\mathbb{E}_{s\sim d_i,\,h\sim q(\cdot\mid s)}
\left[Q_c^i(s,h)\right]
\le \epsilon_{E,i},
\\
&
\mathbb{E}_{s\sim d_i}
D_{\mathrm{KL}}\!\left(
q(\cdot\mid s)\,\|\,\pi_i(\cdot\mid s)
\right)
\le \epsilon_{\mathrm{KL}} .
\end{aligned}
\end{equation}
Here, $\epsilon_{E,i}$ bounds the fixed-reference surrogate cost and is
distinct from the global CMDP limit $d$.

\begin{assumption}[E-step regularity]
\label{ass:estep_regularity}
There exist finite constants $M_r,M_c>0$ such that,
for $d_i(ds)\pi_i(dh\mid s)$-a.e. $(s,h)$,
$|Q_r^i(s,h)|\le M_r$ and $0\le Q_c^i(s,h)\le M_c$. Moreover, there exists a
measurable stochastic kernel $\bar q_i$ from $\mathcal S$ to $\mathcal H$
such that $\bar q_i(\cdot\mid s)\ll\pi_i(\cdot\mid s)$ for $d_i$-a.e. $s$,
whose expected cost is strictly below $\epsilon_{E,i}$ and whose
averaged KL divergence from $\pi_i$ is strictly below
$\epsilon_{\mathrm{KL}}$.
\end{assumption}

The associated Lagrangian, with cost multiplier $\lambda\ge0$ and KL
multiplier $\eta\ge0$, is
\begin{equation}
\label{eq:hcvpo_lagrangian}
\begin{aligned}
\mathcal L_i(q,\eta,\lambda)
=&\;
\mathbb{E}_{s\sim d_i,\,h\sim q(\cdot\mid s)}
\left[Q_r^i(s,h)-\lambda Q_c^i(s,h)\right]
\\
&-\eta\,
\mathbb{E}_{s\sim d_i}
D_{\mathrm{KL}}\!\left(
q(\cdot\mid s)\,\|\,\pi_i(\cdot\mid s)
\right)
+\lambda\epsilon_{E,i}+\eta\epsilon_{\mathrm{KL}} .
\end{aligned}
\end{equation}
For $\eta>0$ and $\lambda\ge0$, define, for $d_i$-a.e. $s$, the
hybrid partition function
\begin{equation}
\label{eq:hcvpo_partition_general}
Z_i(s;\eta,\lambda)
=
\sum_{m\in\mathcal M}
\int_{\mathcal A_c^m}
\pi_i(m,a_c\mid s)
\exp\!\left(
\frac{Q_r^i(s,m,a_c)-\lambda Q_c^i(s,m,a_c)}{\eta}
\right)
\,da_c .
\end{equation}
The corresponding dual objective
$g_i:(0,\infty)\times[0,\infty)\rightarrow\mathbb R$ is
\begin{equation}
\label{eq:hcvpo_dual}
g_i(\eta,\lambda)
=
\eta\epsilon_{\mathrm{KL}}
+\lambda\epsilon_{E,i}
+\eta\,
\mathbb{E}_{s\sim d_i}
\left[\log Z_i(s;\eta,\lambda)\right].
\end{equation}

\begin{theorem}[Strong duality and closed-form hybrid E-step]
\label{thm:hybrid_teacher}
Under Assumption~\ref{ass:estep_regularity}, the E-step has zero duality gap,
and $g_i$ is jointly convex on $(0,\infty)\times[0,\infty)$. If $g_i$ attains
its infimum over this domain at $(\eta_i^\star,\lambda_i^\star)$ with
$\eta_i^\star>0$, then an optimal variational target, defined for
$d_i$-a.e. $s$, has the $\mu$-density
\begin{equation}
\label{eq:hcvpo_teacher}
q_i^\star(m,a_c\mid s)
=
\frac{
\pi_i(m,a_c\mid s)
\exp\!\left(
\frac{
Q_r^i(s,m,a_c)
-\lambda_i^\star Q_c^i(s,m,a_c)
}{
\eta_i^\star
}
\right)
}{
Z_i(s;\eta_i^\star,\lambda_i^\star)
}.
\end{equation}
This target is feasible for Equation~(\ref{eq:hcvpo_estep}) and satisfies
complementary slackness with $(\eta_i^\star,\lambda_i^\star)$.
\end{theorem}

The reward term favors hybrid actions with larger $Q_r^i$, while the cost term
discounts those with larger $Q_c^i$ according to $\lambda_i^\star$. The
E-step therefore reallocates probability jointly over modes and their
conditional controls before projection.

\subsection{Hybrid-Action KL Geometry and Variational Projection}
\label{sec:hybrid_kl}

For a fixed state $s$, write
$q(m,a_c\mid s)=q_m(m\mid s)q_c(a_c\mid s,m)$ and
$\pi(m,a_c\mid s)=\pi_m(m\mid s)\pi_c(a_c\mid s,m)$, where $q_m,\pi_m$ are mode
marginals and $q_c,\pi_c$ are mode-conditioned continuous densities.
If $q(\cdot\mid s)\ll\pi(\cdot\mid s)$, the KL chain rule gives
\begin{equation}
\label{eq:hybrid_kl_decomposition}
\begin{aligned}
D_{\mathrm{KL}}\!\left(q(\cdot\mid s)\,\|\,\pi(\cdot\mid s)\right)
={}&
D_{\mathrm{KL}}\!\left(
q_m(\cdot\mid s)\,\|\,\pi_m(\cdot\mid s)
\right)
\\
&+
\mathbb{E}_{m\sim q_m(\cdot\mid s)}
D_{\mathrm{KL}}\!\left(
q_c(\cdot\mid s,m)\,\|\,\pi_c(\cdot\mid s,m)
\right).
\end{aligned}
\end{equation}
Under the corresponding absolute-continuity conditions, the decomposition
applies to the target-to-policy pair $(q_i^\star,\pi_\theta)$ and the
old-to-new pair $(\pi_i,\pi_\theta)$. Let $\pi_{i,m},\pi_{i,c}$ be the factors
of $\pi_i$, and let $\epsilon_m,\epsilon_c>0$ be the mode and
conditional-continuous trust-region radii. Assuming finite log-density
expectations, the M-step solves
\begin{equation}
\label{eq:hcvpo_mstep}
\begin{aligned}
\max_{\theta}\quad&
\mathbb{E}_{s\sim d_i,\,h\sim q_i^\star(\cdot\mid s)}
\left[\log\pi_\theta(h\mid s)\right]
\\
\mathrm{s.t.}\quad&
\mathbb{E}_{s\sim d_i}
D_{\mathrm{KL}}\!\left(
\pi_{i,m}(\cdot\mid s)\,\|\,\pi_{\theta_m}(\cdot\mid s)
\right)
\le \epsilon_m,
\\
&
\mathbb{E}_{s\sim d_i,\,m\sim \pi_{i,m}(\cdot\mid s)}
D_{\mathrm{KL}}\!\left(
\pi_{i,c}(\cdot\mid s,m)\,\|\,\pi_{\theta_c}(\cdot\mid s,m)
\right)
\le \epsilon_c .
\end{aligned}
\end{equation}

Let $q_{i,m}^\star$ be the mode marginal of $q_i^\star$. Using the policy
factorization in Equation~(\ref{eq:hybrid_policy_pre}), the projection
objective decomposes as
\begin{equation}
\label{eq:factorized_projection}
\begin{aligned}
&\mathbb{E}_{s\sim d_i,\,(m,a_c)\sim q_i^\star(\cdot\mid s)}
\left[\log\pi_\theta(m,a_c\mid s)\right]
\\
={}&
\mathbb{E}_{s\sim d_i,\,m\sim q_{i,m}^\star(\cdot\mid s)}
\left[\log\pi_{\theta_m}(m\mid s)\right]
\\
&+
\mathbb{E}_{s\sim d_i,\,(m,a_c)\sim q_i^\star(\cdot\mid s)}
\left[\log\pi_{\theta_c}(a_c\mid s,m)\right].
\end{aligned}
\end{equation}
Thus, both policy factors are fitted to the same joint variational target.

\subsection{Bounded Approximate Dual and Target Admissibility}
\label{sec:dual_stabilization}

For approximate value functions and expectations, restrict the dual variables
to
\begin{equation}
\label{eq:bounded_dual_domain}
(\eta,\lambda)\in
\mathcal D_i
=
[\eta_{\min},\eta_{\max}]
\times[0,\lambda_{\max}],
\end{equation}
where $0<\eta_{\min}\le\eta_{\max}<\infty$ and
$0<\lambda_{\max}<\infty$.
Let $\widehat g_i:\mathcal D_i\to\mathbb R$ be a continuous approximation of
$g_i$. By compactness, it attains a minimizer
$(\widehat\eta_i,\widehat\lambda_i)\in\mathcal D_i$;
Appendix~\ref{app:proof_bounded_dual} proves this existence claim.

For $(\eta,\lambda)\in\mathcal D_i$, let the normalized approximate E-step
kernel $\widehat q_{i,\eta,\lambda}$ satisfy
$\widehat q_{i,\eta,\lambda}(\cdot\mid s)\ll\pi_i(\cdot\mid s)$ for
$d_i$-a.e. $s$, and set
$\widehat q_i=\widehat q_{i,\widehat\eta_i,\widehat\lambda_i}$.
For the fixed approximate cost action-value function
$\widehat Q_c^i:\mathcal S\times\mathcal H\to\mathbb R$, define the violation
residuals, using $[x]_+=\max\{x,0\}$, by
\begin{equation}
\label{eq:approximate_residuals}
\begin{aligned}
r_{c,i}(q)
&=
\left[
\mathbb{E}_{s\sim d_i,\,h\sim q(\cdot\mid s)}
\left[\widehat Q_c^i(s,h)\right]
-\epsilon_{E,i}
\right]_+,
\\
r_{\mathrm{KL},i}(q)
&=
\left[
\mathbb{E}_{s\sim d_i}
D_{\mathrm{KL}}\!\left(
q(\cdot\mid s)\,\|\,\pi_i(\cdot\mid s)
\right)
-\epsilon_{\mathrm{KL}}
\right]_+ .
\end{aligned}
\end{equation}
For prescribed cost- and KL-residual tolerances
$\tau_c,\tau_{\mathrm{KL}}\ge0$, let
$\mathcal Q_{i,\mathrm{valid}}$ comprise kernels $q$ with finite cost and KL
expectations, $q(\cdot\mid s)\ll\pi_i(\cdot\mid s)$ for $d_i$-a.e. $s$, and
$r_{c,i}(q)\le\tau_c$ and
$r_{\mathrm{KL},i}(q)\le\tau_{\mathrm{KL}}$. Run the M-step only if
$\widehat q_i\in\mathcal Q_{i,\mathrm{valid}}$; otherwise
$\pi_{i+1}=\pi_i$.

\subsection{Theoretical Properties}
\label{sec:theory_properties}

For a positive-temperature optimal E-step dual pair
$(\eta_i^\star,\lambda_i^\star)$ satisfying
Theorem~\ref{thm:hybrid_teacher}, define, for a stochastic kernel $q$ and
hybrid policy $\pi$, the fixed-reference surrogate
\begin{equation}
\label{eq:fixed_reference_surrogate}
\begin{aligned}
\mathcal F_i(q,\pi)
={}&
\mathbb{E}_{s\sim d_i,\,h\sim q(\cdot\mid s)}
\left[
Q_r^i(s,h)-\lambda_i^\star Q_c^i(s,h)
\right]
\\
&-
\eta_i^\star
\mathbb{E}_{s\sim d_i}
D_{\mathrm{KL}}\!\left(
q(\cdot\mid s)\,\|\,\pi(\cdot\mid s)
\right).
\end{aligned}
\end{equation}

\begin{proposition}[Fixed-reference E/M surrogate improvement]
\label{prop:fixed_reference_improvement}
Assume exact E/M solutions and finiteness of the displayed quantities. Then,
for every stochastic kernel $q$ with
$q(\cdot\mid s)\ll\pi_i(\cdot\mid s)$ for $d_i$-a.e. $s$ and finite
$\mathcal F_i(q,\pi_i)$,
\begin{equation}
\mathcal F_i(q_i^\star,\pi_{i+1})
\ge
\mathcal F_i(q_i^\star,\pi_i)
\ge
\mathcal F_i(q,\pi_i).
\end{equation}
\end{proposition}

The inequalities are the exact M- and E-step improvements, respectively, with
$d_i,Q_r^i,Q_c^i,\eta_i^\star$, and $\lambda_i^\star$ held fixed.

The outer CMDP analysis instead uses the true normalized discounted occupancy
measure of $\pi_i$,
\begin{equation}
d_{\pi_i}
=
(1-\gamma)
\sum_{t=0}^{\infty}
\gamma^t
\Pr_{\pi_i}(s_t\in\cdot).
\end{equation}
For $x\in\{r,c\}$, let $Q_x^{\pi_i}$ and $V_x^{\pi_i}$ be the true discounted
action- and state-value functions, and define
$A_x^{\pi_i}(s,h)=Q_x^{\pi_i}(s,h)-V_x^{\pi_i}(s)$. Define
\begin{equation}
\Delta_i
=
\sup_{s\in\mathcal S}
D_{\mathrm{TV}}\!\left(
\pi_{i+1}(\cdot\mid s),\pi_i(\cdot\mid s)
\right),
\end{equation}
where, for probability measures $P,Q$ on a common measurable space,
$D_{\mathrm{TV}}(P,Q)=\sup_C|P(C)-Q(C)|$, with $C$ ranging over measurable
sets. Define also
\begin{equation}
B_{x,i+1}
=
\sup_{s\in\mathcal S}
\left|
\mathbb{E}_{h\sim\pi_{i+1}(\cdot\mid s)}
\left[A_x^{\pi_i}(s,h)\right]
\right|,
\qquad x\in\{r,c\}.
\end{equation}

The following proposition adapts the performance bounds of
\citet{Achiam2017CPO}.

\begin{proposition}[Conditional reward and cost bounds]
\label{prop:cost_variation}
Assume $B_{r,i+1}$ and $B_{c,i+1}$ are finite. Then
\begin{equation}
\begin{aligned}
J_r(\pi_{i+1})
\ge{}&
J_r(\pi_i)
+
\frac{1}{1-\gamma}
\mathbb{E}_{\substack{s\sim d_{\pi_i}\\h\sim\pi_{i+1}(\cdot\mid s)}}
\left[A_r^{\pi_i}(s,h)\right]
-
\frac{2\gamma B_{r,i+1}}{(1-\gamma)^2}\Delta_i,
\\
J_c(\pi_{i+1})
\le{}&
J_c(\pi_i)
+
\frac{1}{1-\gamma}
\mathbb{E}_{\substack{s\sim d_{\pi_i}\\h\sim\pi_{i+1}(\cdot\mid s)}}
\left[A_c^{\pi_i}(s,h)\right]
+
\frac{2\gamma B_{c,i+1}}{(1-\gamma)^2}\Delta_i .
\end{aligned}
\end{equation}
\end{proposition}

Define the cost-increase upper bound
\begin{equation}
U_{c,i}
=
\frac{1}{1-\gamma}
\mathbb{E}_{\substack{s\sim d_{\pi_i}\\h\sim\pi_{i+1}(\cdot\mid s)}}
\left[A_c^{\pi_i}(s,h)\right]
+
\frac{2\gamma B_{c,i+1}}{(1-\gamma)^2}\Delta_i .
\end{equation}

If $J_c(\pi_i)\le d-\zeta$ for some constraint margin $\zeta>0$ and
$U_{c,i}\le\zeta$, then $J_c(\pi_{i+1})\le d$, giving a sufficient
feasibility margin condition.

Proposition~\ref{prop:fixed_reference_improvement} establishes exact
fixed-reference surrogate improvement. Proposition~\ref{prop:cost_variation}
and the margin condition characterize the outer CMDP update using true
occupancies and advantages. These bounds are conditional on the statewise
shift $\Delta_i$, which is not controlled by the averaged M-step KL
constraints.

\section{Experiments}
\label{sec:experiments}

We compare H-CVPO with constrained hybrid-action baselines at three cost
limits under matched interaction budgets and conduct component-wise
ablations. The main text presents the experimental setup, baseline
comparisons, and main ablation results. Appendix~\ref{app:experimental_details}
provides training and task details, learning curves at additional cost
limits, late-stage trends, an extended-training comparison with CHPO,
and complete ablation results.

\subsection{Experimental Setup}
\label{sec:tasks_baselines}

\textbf{Tasks and baselines.}
To support fair and reproducible comparisons, we follow
CHPO~\citep{Zhou2025CHPO} in evaluating constrained versions of
\emph{Moving}, \emph{Sliding}, \emph{HardMove}, and \emph{Parking}, which
respectively test mode--parameter coordination, inertial dynamics, combinatorial
mode spaces, and collision-aware maneuvering. We compare H-CVPO with CHPO,
HPPO-Lag, and PDQN-Rco: CHPO switches between reward and cost objectives, while
HPPO-Lag and PDQN-Rco use Lagrangian and reward-constrained updates.
The public CHPO implementation is available in its source-code repository
~\citep{Zhou2025CHPOCode}.
Appendix~\ref{app:tasks} provides task details
and clarifies the scope of the theoretical and empirical comparisons.

\textbf{Protocol.}
Unless stated otherwise, all experiments use training seeds 11, 22, and 33,
with matched environments, interaction budgets, cost limits, and evaluation
protocols across methods. For the standard 320k experiments, all tabulated
final-policy results use the policy obtained at the fixed training endpoint
for every method and seed. We do not select checkpoints after training,
keeping both the training budget and checkpoint-selection rule matched
across methods.

\begin{figure*}[htbp]
    \centering
    \includegraphics[width=\textwidth]
    {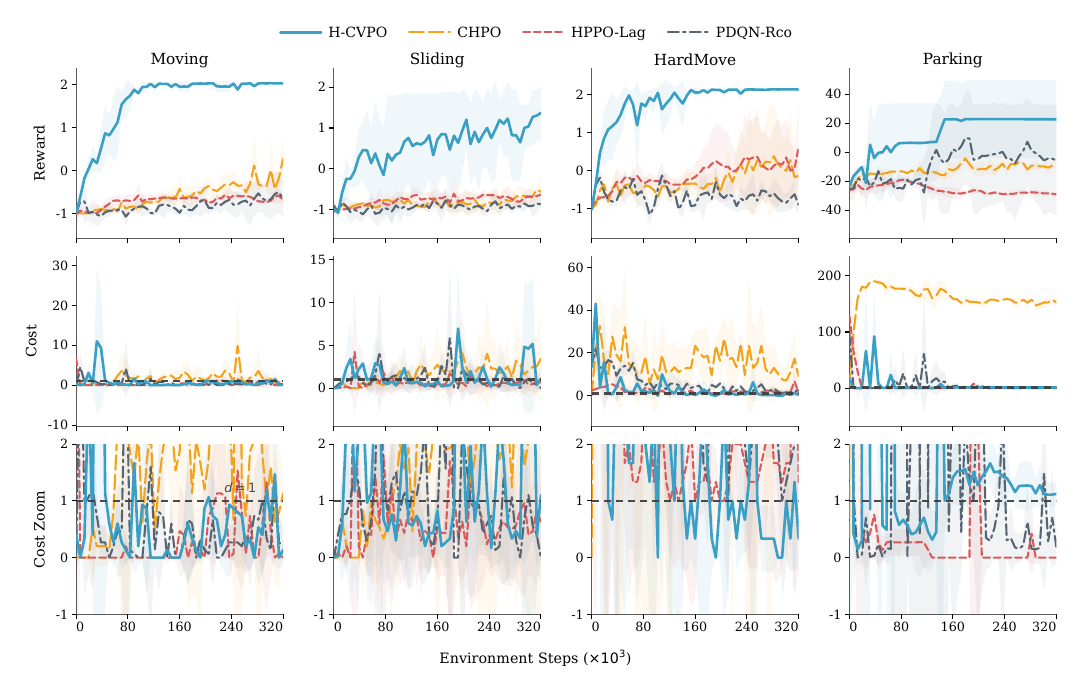}
    \caption{
    Learning curves for H-CVPO and the baselines under the primary cost limit
    $d=1.0$. Lines and shaded regions show the mean and one standard deviation
    over the three training seeds. Rows report reward, cost on the full scale,
    and cost near the constraint boundary; horizontal dashed lines mark the
    cost limit.
    }
    \label{fig:main_comparison}
\end{figure*}

\subsection{Comparison with Baselines}
\label{sec:performance_comparison}

\paragraph{Performance across hybrid-control tasks.}
Figure~\ref{fig:main_comparison} compares H-CVPO with the baselines under the
primary cost limit $d=1.0$. By roughly $120$k environment steps, mean rewards
on Moving and HardMove approach $2$ and subsequently stabilize, while their
pronounced initial cost spikes largely subside. Parking undergoes a second
stage of improvement around $140$--$160$k steps, when reward rises from its
intermediate plateau to above $20$; mean cost is below the limit at several
earlier checkpoints but rises above it during this reward increase. Sliding
improves more gradually, with reward fluctuations and recurrent cost spikes
throughout the $320$k-step budget.

CHPO shows modest late-stage reward gains on Moving and HardMove, but continues
to incur substantial costs on HardMove and especially Parking through $320$k
steps. HPPO-Lag and PDQN-Rco generally maintain lower costs, but achieve
substantially lower rewards than H-CVPO on Moving, Sliding, and HardMove.
On Parking, both baselines also achieve substantially lower late-stage
rewards than H-CVPO.

Feasibility requires mean episodic evaluation cost not exceeding $d$.
Evaluations during training identify multiple feasible H-CVPO checkpoints
per task (Figure~\ref{fig:main_comparison}).
Following Section~\ref{sec:tasks_baselines}, we compare only
$320$k-endpoint policies in Table~\ref{tab:final40_main_neutral},
with matched training budgets and uniform checkpoint-selection and
evaluation protocols, without post-hoc selection of earlier feasible
checkpoints.

\begin{table*}[t]
\centering
\scriptsize
\setlength{\tabcolsep}{2.4pt}
\caption{Final performance after 320k environment steps under different cost
limits. For each training seed, we average 40 deterministic evaluation episodes
using fixed evaluation seeds 22000--22039 and report the mean and sample
standard deviation across the three training seeds.}
\label{tab:final40_main_neutral}
\begin{tabular}{@{}ll*{8}{c}@{}}
\toprule
& & \multicolumn{2}{c}{H-CVPO} & \multicolumn{2}{c}{CHPO} & \multicolumn{2}{c}{HPPO-Lag} & \multicolumn{2}{c}{PDQN-Rco} \\
\cmidrule(lr){3-4}\cmidrule(lr){5-6}\cmidrule(lr){7-8}\cmidrule(l){9-10}
Environment & $d$ & Reward & Cost & Reward & Cost & Reward & Cost & Reward & Cost \\
\midrule
\multirow{3}{*}{Moving} & 1.0 & 2.07$\pm$0.03 & 0.85$\pm$0.97 & -0.04$\pm$0.19 & 0.98$\pm$0.73 & -0.69$\pm$0.09 & 1.30$\pm$1.51 & -0.72$\pm$0.43 & 0.12$\pm$0.20 \\
 & 1.5 & 1.41$\pm$1.18 & 1.07$\pm$0.60 & -0.12$\pm$0.35 & 1.83$\pm$0.52 & -0.62$\pm$0.08 & 1.33$\pm$1.50 & -0.60$\pm$0.49 & 0.28$\pm$0.30 \\
 & 2.0 & 2.08$\pm$0.02 & 1.00$\pm$0.43 & -0.12$\pm$0.35 & 1.83$\pm$0.52 & -0.46$\pm$0.10 & 0.37$\pm$0.16 & 0.02$\pm$0.43 & 5.83$\pm$7.87 \\
\midrule
\addlinespace[2pt]
\multirow{3}{*}{Sliding} & 1.0 & 1.08$\pm$0.61 & 0.67$\pm$0.47 & -0.63$\pm$0.21 & 3.79$\pm$1.71 & -0.67$\pm$0.16 & 0.61$\pm$0.38 & -0.95$\pm$0.05 & 0.49$\pm$0.48 \\
 & 1.5 & 1.55$\pm$0.85 & 0.68$\pm$0.30 & -0.62$\pm$0.26 & 2.06$\pm$0.55 & -0.64$\pm$0.04 & 1.07$\pm$1.24 & -0.43$\pm$0.38 & 2.19$\pm$0.83 \\
 & 2.0 & 1.99$\pm$0.10 & 0.84$\pm$0.06 & -0.79$\pm$0.15 & 2.08$\pm$0.46 & -0.64$\pm$0.09 & 0.46$\pm$0.14 & -0.63$\pm$0.10 & 3.52$\pm$5.55 \\
\midrule
\addlinespace[2pt]
\multirow{3}{*}{HardMove} & 1.0 & 2.05$\pm$0.00 & 0.42$\pm$0.52 & -0.33$\pm$0.43 & 10.58$\pm$4.93 & 0.34$\pm$1.25 & 1.17$\pm$0.38 & -0.73$\pm$0.12 & 2.25$\pm$2.17 \\
 & 1.5 & 2.06$\pm$0.01 & 2.17$\pm$3.32 & -0.29$\pm$0.39 & 9.00$\pm$2.22 & -0.31$\pm$0.22 & 3.25$\pm$2.61 & -0.31$\pm$0.39 & 2.00$\pm$2.38 \\
 & 2.0 & 2.05$\pm$0.02 & 0.50$\pm$0.43 & -0.31$\pm$0.42 & 13.00$\pm$6.01 & -0.33$\pm$0.18 & 1.67$\pm$0.38 & -0.85$\pm$0.15 & 2.33$\pm$0.88 \\
\midrule
\addlinespace[2pt]
\multirow{3}{*}{Parking} & 1.0 & 23.26$\pm$26.94 & 1.06$\pm$0.16 & 7.45$\pm$3.31 & 91.98$\pm$3.59 & -23.78$\pm$2.12 & 0.15$\pm$0.05 & 0.37$\pm$33.49 & 0.19$\pm$0.29 \\
 & 1.5 & 38.82$\pm$0.01 & 0.85$\pm$0.28 & 7.45$\pm$3.31 & 91.98$\pm$3.59 & -23.36$\pm$0.35 & 0.12$\pm$0.00 & -14.60$\pm$9.24 & 63.83$\pm$104.30 \\
 & 2.0 & 35.92$\pm$5.03 & 1.23$\pm$0.13 & 7.45$\pm$3.31 & 91.98$\pm$3.59 & -22.54$\pm$0.59 & 0.12$\pm$0.02 & -22.86$\pm$9.95 & 59.64$\pm$49.87 \\
\bottomrule
\end{tabular}
\end{table*}

\paragraph{Performance under different cost limits.}
Table~\ref{tab:final40_main_neutral} reports final-policy performance across
12 task--limit settings with $d\in\{1.0,1.5,2.0\}$. Under the evaluation-based
feasibility criterion above, H-CVPO satisfies the limit in 10 settings and
achieves the highest mean reward in all 12. Its two endpoint cost violations
occur on Parking at $d=1.0$, where the mean cost is $1.06$, and on HardMove
at $d=1.5$, where the mean cost is $2.17$. No evaluated method is feasible in
the latter setting.

CHPO is feasible only on Moving at $d=1.0$ and $2.0$, with costs of
$9.00$--$13.00$ on HardMove and $91.98$ on Parking. Its Sliding cost of
$2.08$ at $d=2.0$, like H-CVPO's Parking cost of $1.06$ at $d=1.0$,
counts as infeasible despite the small exceedance. HPPO-Lag is feasible in
9 settings but has lower mean rewards than H-CVPO in all 12, often with
costs well below the limits. PDQN-Rco is feasible in 4 settings, with large
cost exceedances and cross-seed variability in several cases. Thus, H-CVPO
has both the highest feasibility count and the highest mean rewards.

An additional observation concerns the identical CHPO statistics reported under different cost limits. In both the final-policy results (Table~\ref{tab:final40_main_neutral}) and the late-stage averages (Table~\ref{tab:late5_appendix_neutral}), CHPO reports identical reward and cost means and standard deviations on Moving at $d=1.5$ and $2.0$, and on Parking across all three cost limits. This behavior can be understood from CHPO's feasibility-dependent objective switching. Specifically, CHPO selects cost minimization when its estimated undiscounted cost return exceeds the prescribed limit, and reward maximization otherwise. For example, when the estimated cost return exceeds $2.0$, both $d=1.5$ and $d=2.0$ select the cost-minimization branch. Consequently, different cost limits may induce the same optimization branch along a training trajectory, leading to identical policy updates and reported statistics. Further investigation of CHPO's internal update dynamics is beyond the scope of this work, which focuses on H-CVPO.

\subsection{Ablation Studies}
\label{sec:ablation}

We examine three single-component variants under $d=1.0$.
\emph{w/o Cost-Aware E-Step} removes only the cost term from the joint target in
Equation~(\ref{eq:hcvpo_teacher}); the E-step itself, the cost critic,
and all other update components are retained. \emph{Joint Hybrid KL} replaces the two old-to-new M-step
KL constraints in Equation~(\ref{eq:hcvpo_mstep}) with a single joint KL
constraint at the same total KL allowance. \emph{w/o Target Admissibility}
removes only the pre-projection filter based on the cost and KL residuals in
Equation~(\ref{eq:approximate_residuals}), disabling the \emph{Skip} branch
in Algorithm~\ref{alg:hcvpo_training} while retaining post-update acceptance
checks. All other training and evaluation settings remain fixed. We focus
on the cost-aware E-step here; Appendix~\ref{app:ablation_studies} reports
all three variants across $d\in\{1.0,1.5,2.0\}$.

\begin{figure*}[htbp]
    \centering
    \includegraphics[width=\textwidth]
    {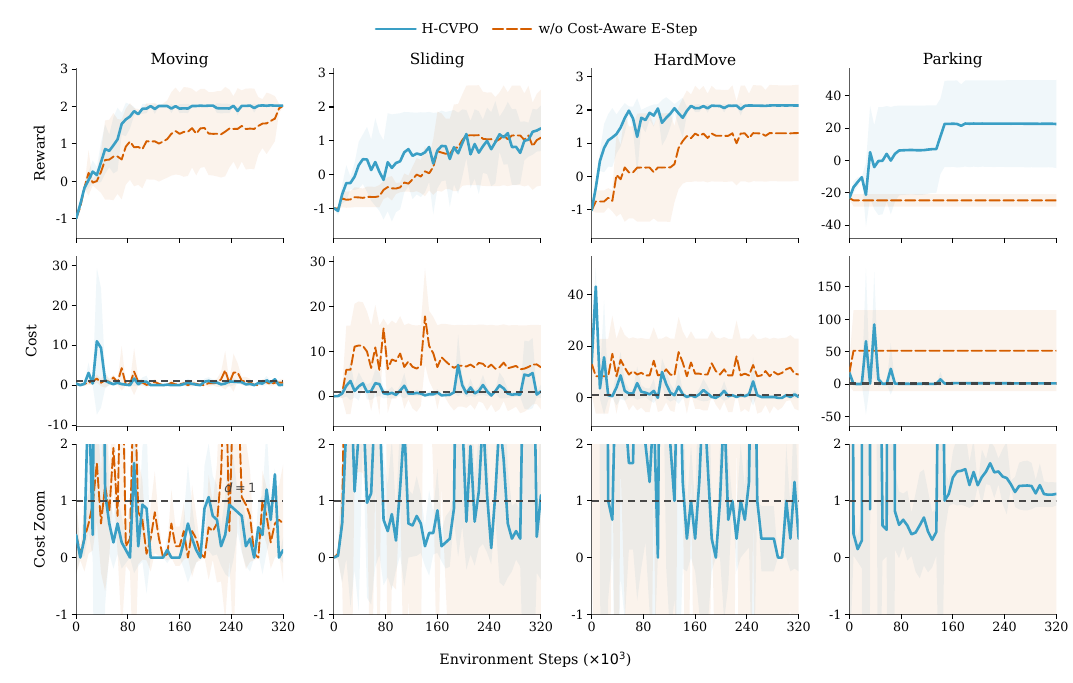}
    \caption{
    Ablation of the cost-aware variational E-step under the primary cost
    limit $d=1.0$. Curves compare full H-CVPO with the variant removing
    the cost contribution from the variational target. Curves and shading
    denote the mean and one standard deviation over the three training
    seeds; dashed lines mark the cost limit.
    }
    \label{fig:ablation}
\end{figure*}

Figure~\ref{fig:ablation} compares full H-CVPO with the reward-only E-step
variant. The ablation's mean cost exceeds $d=1.0$ for most of training on
Sliding, HardMove, and Parking. Full H-CVPO has substantially lower costs on
these tasks, although Sliding and HardMove retain cost spikes and Parking
remains above the limit late in training. Removing cost awareness
also delays reward improvement on Moving and yields lower rewards on
HardMove and Parking; Sliding has broadly comparable rewards later in
training.

The fixed $320$k-endpoint evaluations in
Table~\ref{tab:hcvpo_ablation_cross_d_final_policy} provide a complementary
comparison. At $d=1.0$, removing cost awareness changes Moving and HardMove
from feasible to infeasible, raising mean costs from $0.85$ to $1.73$ and
from $0.42$ to $3.25$, respectively. On Parking, full H-CVPO already slightly
exceeds the limit; the ablation lowers mean reward from $23.26$ to $-18.30$
and raises mean cost from $1.06$ to $51.39$. Sliding is an exception: both
endpoint policies are feasible, and the ablation raises mean reward from
$1.08$ to $1.31$ while lowering mean cost from $0.67$ to $0.52$, but increases
the across-seed reward standard deviation from $0.61$ to $1.32$.
Across all three cost limits, removing cost-aware reweighting changes eight
of the twelve task--limit settings from feasible to infeasible. These
results support cost-aware variational reweighting as an important
contributor to constraint control, with task- and limit-dependent exceptions.

\FloatBarrier

\section{Conclusion}

We develop H-CVPO to address constrained reinforcement learning in hybrid
action spaces, coordinating mode selection and mode-conditioned control under
long-horizon cost constraints. Its E-step forms a reward--cost-aware joint
nonparametric target, while its M-step fits both policy components under
separate discrete and conditional-continuous KL controls. Under the stated assumptions,
exact E/M updates yield fixed-reference surrogate improvement and conditional
reward--cost bounds; approximate updates use cost and KL residuals to determine
target projection. Across four tasks and three cost limits, H-CVPO shows favorable
reward--cost trade-offs against hybrid-action baselines. Ablations identify
cost-aware reweighting as important for constraint control, while the effects of
separate KL controls and target-admissibility filtering vary by task and cost
limit. As discrete--continuous decision structures arise broadly in engineered
systems, we expect H-CVPO to offer a general approach to constrained hybrid
control across industrial systems and beyond.

\clearpage
{\sloppy
\setlength{\bibsep}{8pt plus 4pt minus 2pt}
\bibliography{iclr2027_hcvpo_refs}
\bibliographystyle{iclr2027_conference}
}

\captionsetup{font=default,skip=10pt,hypcap=false}
\setlength{\textfloatsep}{20pt plus 2pt minus 4pt}
\setlength{\floatsep}{12pt plus 2pt minus 2pt}
\setlength{\intextsep}{12pt plus 2pt minus 2pt}
\setlength{\abovecaptionskip}{10pt}
\setlength{\belowcaptionskip}{0pt}
\makeatletter
\setlength{\@fpsep}{8pt plus 2fil}
\makeatother

\appendix
\section{Proofs and Derivations}
\label{app:proofs}

We provide full proofs and derivations of the results in
Section~\ref{sec:hcvpo}.

\subsection{Proof of Strong Duality and Closed-Form Hybrid E-Step}
\label{app:proof_teacher}

We retain the notation of Section~\ref{sec:mixed_estep}, including the
reference measure $d_i$, current policy $\pi_i$, action-value functions
$Q_r^i,Q_c^i$, hybrid reference measure $\mu$, and E-step bounds
$\epsilon_{E,i},\epsilon_{\mathrm{KL}}$.

\paragraph{Density-ratio formulation.}
Define the joint probability measure $\nu_i$ on
$\mathcal S\times\mathcal H$ by
\begin{equation}
\nu_i(ds,dh)=d_i(ds)\pi_i(dh\mid s).
\end{equation}
For an admissible kernel $q$, write $q(h\mid s)$ and $\pi_i(h\mid s)$ for
their $\mu$-densities. Absolute continuity gives the density ratio
\begin{equation}
w(s,h)=\frac{q(h\mid s)}{\pi_i(h\mid s)},
\end{equation}
defined $\nu_i$-a.e. Let $L^1(\nu_i)$ denote the space of
$\nu_i$-integrable functions identified up to $\nu_i$-a.e. equality, with
norm $\|w\|_{L^1(\nu_i)}=\int|w|\,d\nu_i$; define $L^1(d_i)$ analogously.
Let $\mathcal W_i$ denote the admissible density ratios:
\begin{equation}
\mathcal W_i
=
\left\{
w\in L^1(\nu_i):
w\ge0,\quad
\int_{\mathcal H}w(s,h)\pi_i(dh\mid s)=1
\ \text{for }d_i\text{-a.e. }s
\right\}.
\end{equation}
For each $w\in\mathcal W_i$, choose a jointly measurable nonnegative
representative and set $q(h\mid s)=w(s,h)\pi_i(h\mid s)$ on the states
where the normalization holds. On the exceptional measurable $d_i$-null
set, set $q(\cdot\mid s)=\pi_i(\cdot\mid s)$. This defines an admissible
kernel without changing any $\nu_i$-integral. We use the same null-set
convention for subsequent kernel constructions. The conditional-integration
map is bounded and linear from $L^1(\nu_i)$ to $L^1(d_i)$. Together with
the weak closedness of the nonnegative cone, this shows that
$\mathcal W_i$ is weakly closed and convex.

For $w\in\mathcal W_i$, define
\begin{equation}
\begin{aligned}
R_i(w)
&=
\int_{\mathcal S\times\mathcal H}
w(s,h)Q_r^i(s,h)\,\nu_i(ds,dh),\\
C_i(w)
&=
\int_{\mathcal S\times\mathcal H}
w(s,h)Q_c^i(s,h)\,\nu_i(ds,dh),\\
H_i(w)
&=
\int_{\mathcal S\times\mathcal H}
w(s,h)\log w(s,h)\,\nu_i(ds,dh),
\end{aligned}
\end{equation}
with $0\log0=0$. Here, $R_i(w)$ and $C_i(w)$ are the fixed-reference
expectations of the reward and cost action-values, while $H_i(w)$ is the
averaged KL divergence in Equation~(\ref{eq:hcvpo_estep}).

\paragraph{Primal attainment.}
Assumption~\ref{ass:estep_regularity} makes $R_i$ and $C_i$ weakly continuous
linear functionals on $L^1(\nu_i)$. The entropy functional $H_i$ is proper,
convex, and weakly lower semicontinuous. Every $w\in\mathcal W_i$ satisfies
$\int w\,d\nu_i=1$. For a finite entropy bound $K\ge0$ and a tail
threshold $t>1$,
\begin{equation}
\sup_{\substack{w\in\mathcal W_i\\H_i(w)\le K}}
\int_{\{w>t\}}w\,d\nu_i
\le \frac{K+1/e}{\log t},
\end{equation}
since $w\log w\ge-1/e$. Thus, these sublevel sets are uniformly
integrable and hence relatively weakly compact by the Dunford--Pettis
theorem. They are also weakly closed, by the weak lower semicontinuity of
$H_i$ and weak closedness of $\mathcal W_i$, and are therefore weakly
compact. Let $p_i^\star$ denote the optimal E-step value:
\begin{equation}
\label{eq:app_estep_value}
\begin{aligned}
p_i^\star
=
\sup_{w\in\mathcal W_i}\quad &R_i(w)\\
\mathrm{s.t.}\quad
&C_i(w)\le\epsilon_{E,i},\qquad
H_i(w)\le\epsilon_{\mathrm{KL}}
\end{aligned}
\end{equation}
The feasible set is nonempty by strict feasibility and weakly compact by
the preceding argument and weak continuity of $C_i$. Hence the weakly
continuous objective $R_i$ attains its supremum.

\paragraph{Strong duality.}
To establish strong duality, let $u_c$ and $u_{\mathrm{KL}}$ denote
perturbed upper bounds on cost and KL divergence, respectively, and let
$u_r$ denote a perturbed reward level. Define
\begin{equation}
\mathcal V_i
=
\left\{
(u_c,u_{\mathrm{KL}},u_r)\in\mathbb R^3:
\begin{array}{l}
\text{there exists }w\in\mathcal W_i\text{ such that}\\
C_i(w)\le u_c,\ H_i(w)\le u_{\mathrm{KL}},\ R_i(w)\ge u_r
\end{array}
\right\}.
\end{equation}
Convexity of $H_i$ makes $\mathcal V_i$ convex. To see that it is closed,
consider a convergent sequence in $\mathcal V_i$. Its entropy coordinates
are bounded, so the corresponding witnesses lie in a common weakly compact
entropy sublevel set. Passing to a weakly convergent subsequence gives a
limit in $\mathcal W_i$; weak continuity of $R_i,C_i$ and weak lower
semicontinuity of $H_i$ preserve the three defining inequalities.

The point
$(\epsilon_{E,i},\epsilon_{\mathrm{KL}},p_i^\star)$ lies on the boundary of
$\mathcal V_i$. Its monotonicity in the three coordinates implies that a
supporting hyperplane can be written with the cost, KL, and objective
coefficients $\lambda,\eta,\alpha\ge0$, not all zero, as
\begin{equation}
\lambda(u_c-\epsilon_{E,i})
+\eta(u_{\mathrm{KL}}-\epsilon_{\mathrm{KL}})
-\alpha(u_r-p_i^\star)
\ge0,
\qquad
(u_c,u_{\mathrm{KL}},u_r)\in\mathcal V_i.
\end{equation}
If $\alpha=0$, evaluating this inequality at the density ratio associated
with the strictly feasible kernel in Assumption~\ref{ass:estep_regularity}
forces $\lambda=\eta=0$, a contradiction. Thus, $\alpha>0$, and rescaling
gives $\alpha=1$. Consequently,
for every
$w\in\mathcal W_i$ with finite $H_i(w)$,
\begin{equation}
R_i(w)
-\lambda\bigl(C_i(w)-\epsilon_{E,i}\bigr)
-\eta\bigl(H_i(w)-\epsilon_{\mathrm{KL}}\bigr)
\le p_i^\star .
\end{equation}
This inequality extends to the full Lagrangian domain. If $\eta>0$,
ratios with $H_i(w)=\infty$ have Lagrangian value $-\infty$. If $\eta=0$,
the entropy term is omitted, and finite-entropy ratios suffice by the
following approximation. For any $w\in\mathcal W_i$ and integer $n\ge1$,
define the retained conditional mass and a bounded density ratio by
\begin{equation}
b_n(s)
=
\int_{\mathcal H}\min\{w(s,h),n\}\pi_i(dh\mid s),
\qquad
w_n(s,h)
=
\min\{w(s,h),n\}+1-b_n(s).
\end{equation}
Then $w_n\in\mathcal W_i$ and $0\le w_n\le n+1$ almost everywhere.
Moreover,
\begin{equation}
\|w_n-w\|_{L^1(\nu_i)}
\le 2\int\bigl(w-\min\{w,n\}\bigr)\,d\nu_i
\longrightarrow 0.
\end{equation}
Thus, $H_i(w_n)<\infty$, while boundedness of $Q_r^i$ and $Q_c^i$ gives
$R_i(w_n)\to R_i(w)$ and $C_i(w_n)\to C_i(w)$. This extends the
inequality to every $w\in\mathcal W_i$ when $\eta=0$, without requiring
entropy convergence. The dual objective at the supporting multipliers is
therefore at most $p_i^\star$; weak duality gives the reverse inequality.
Hence the E-step has zero duality gap on the full multiplier domain.

\paragraph{The positive-temperature Gibbs target.}
We next identify the Lagrangian maximizer. For fixed $\eta>0$ and
$\lambda\ge0$, define the probability kernel $q_{\eta,\lambda}$, with the
fixed iteration index $i$ suppressed, using the partition function
$Z_i$ from Equation~(\ref{eq:hcvpo_partition_general}). Its $\mu$-density is
\begin{equation}
q_{\eta,\lambda}(h\mid s)
=
\frac{
\pi_i(h\mid s)
\exp\!\left(
\frac{Q_r^i(s,h)-\lambda Q_c^i(s,h)}{\eta}
\right)
}{
Z_i(s;\eta,\lambda)
}.
\end{equation}
The boundedness in Assumption~\ref{ass:estep_regularity} makes the partition
function finite and strictly positive for $d_i$-a.e. $s$. With the null-set
convention above, this defines a measurable probability kernel with
$q_{\eta,\lambda}(\cdot\mid s)\ll\pi_i(\cdot\mid s)$. Its log-density
ratio relative to $\pi_i$ is bounded $\nu_i$-a.e. for fixed
$(\eta,\lambda)$, so its averaged KL divergence is finite. For any admissible
kernel $q$ with finite averaged KL divergence, direct expansion gives, for
$d_i$-a.e. $s$,
\begin{equation}
\begin{aligned}
&\mathbb{E}_{h\sim q(\cdot\mid s)}
\left[Q_r^i(s,h)-\lambda Q_c^i(s,h)\right]
-\eta D_{\mathrm{KL}}\!\left(
q(\cdot\mid s)\,\|\,\pi_i(\cdot\mid s)
\right)
\\
&\qquad=
\eta\log Z_i(s;\eta,\lambda)
-\eta D_{\mathrm{KL}}\!\left(
q(\cdot\mid s)\,\|\,q_{\eta,\lambda}(\cdot\mid s)
\right).
\end{aligned}
\end{equation}
The last term is nonpositive and vanishes precisely at
$q=q_{\eta,\lambda}$ up to null sets. Hence $q_{\eta,\lambda}$ maximizes the
Lagrangian; kernels of infinite averaged KL divergence have Lagrangian value
$-\infty$. Substitution yields the dual objective
$g_i(\eta,\lambda)$ in Equation~(\ref{eq:hcvpo_dual}).

\paragraph{The zero-temperature boundary.}
The multiplier domain of the Lagrange dual includes $\eta=0$, whereas
Equation~(\ref{eq:hcvpo_dual}) is written for $\eta>0$. We verify that its
boundary limit equals the actual Lagrangian supremum with the entropy
term omitted. For fixed $\lambda\ge0$, define
\begin{equation}
S_{i,\lambda}(s)
=
\operatorname*{ess\,sup}_{h\sim\pi_i(\cdot\mid s)}
\left\{Q_r^i(s,h)-\lambda Q_c^i(s,h)\right\}.
\end{equation}
The Laplace principle gives, for $d_i$-a.e. $s$,
\begin{equation}
\eta\log Z_i(s;\eta,\lambda)
\longrightarrow S_{i,\lambda}(s)
\qquad\text{as }\eta\downarrow0.
\end{equation}
This limit provides a measurable version of $S_{i,\lambda}$. Both sides
are bounded in absolute value by $M_r+\lambda M_c$, so dominated
convergence permits integration over $d_i$. Every admissible kernel has
expected reward--cost action-value at most
$\mathbb E_{s\sim d_i}[S_{i,\lambda}(s)]$. Conversely, evaluating the
Gibbs identity at $q_{\eta,\lambda}$ and using nonnegativity of KL gives
\begin{equation}
\begin{aligned}
\eta\mathbb E_{s\sim d_i}[\log Z_i(s;\eta,\lambda)]
&\le
\mathbb E_{s\sim d_i,\,h\sim q_{\eta,\lambda}(\cdot\mid s)}
[Q_r^i(s,h)-\lambda Q_c^i(s,h)]\\
&\le \mathbb E_{s\sim d_i}[S_{i,\lambda}(s)].
\end{aligned}
\end{equation}
Letting $\eta\downarrow0$ proves that these kernels approach the upper
bound. Therefore the boundary dual objective is
\begin{equation}
\begin{aligned}
g_i(0,\lambda)
&=\lambda\epsilon_{E,i}
+\sup_{w\in\mathcal W_i}\{R_i(w)-\lambda C_i(w)\}\\
&=\lambda\epsilon_{E,i}
+\mathbb E_{s\sim d_i}[S_{i,\lambda}(s)]
=\lim_{\eta\downarrow0}g_i(\eta,\lambda).
\end{aligned}
\end{equation}
Thus, including the boundary value $\eta=0$ leaves the dual optimal value
unchanged: it is $\inf_{\eta>0,\,\lambda\ge0}g_i(\eta,\lambda)$. However,
this does not guarantee that an optimizer exists with $\eta>0$.

\paragraph{Convexity and optimality conditions.}
It remains to establish convexity and the optimality conditions. For
$d_i$-a.e. state $s$, let $\widetilde q(\cdot\mid s)$ range over probability measures on
$\mathcal H$ that are absolutely continuous with respect to
$\pi_i(\cdot\mid s)$. The preceding KL identity is equivalently the Gibbs
variational formula
\begin{equation}
\eta\log Z_i(s;\eta,\lambda)
=
\sup_{\widetilde q(\cdot\mid s)\ll\pi_i(\cdot\mid s)}
\left\{
\mathbb{E}_{h\sim\widetilde q(\cdot\mid s)}
\left[Q_r^i(s,h)-\lambda Q_c^i(s,h)\right]
-\eta D_{\mathrm{KL}}\!\left(
\widetilde q(\cdot\mid s)\,\|\,\pi_i(\cdot\mid s)
\right)
\right\}.
\end{equation}
Measures with infinite KL divergence contribute $-\infty$ and may be
excluded from the supremum. For each remaining $\widetilde q$, the
expression in braces is affine in $(\eta,\lambda)$; its supremum is
therefore jointly convex. Averaging over
$d_i$ and adding the affine terms
$\eta\epsilon_{\mathrm{KL}}+\lambda\epsilon_{E,i}$ preserves convexity.
Consequently, $g_i$ is jointly convex on
$(0,\infty)\times[0,\infty)$. Moreover, the boundedness of $Q_r^i$ and
$Q_c^i$ supplies local dominating functions, so differentiation may pass
through the state and action integrals for $\eta>0$. The Gibbs kernel gives
\begin{equation}
\begin{aligned}
\frac{\partial g_i}{\partial\eta}
&=
\epsilon_{\mathrm{KL}}
-\mathbb{E}_{s\sim d_i}
D_{\mathrm{KL}}\!\left(
q_{\eta,\lambda}(\cdot\mid s)\,\|\,\pi_i(\cdot\mid s)
\right),
\\
\frac{\partial g_i}{\partial\lambda}
&=
\epsilon_{E,i}
-\mathbb{E}_{s\sim d_i,\,h\sim q_{\eta,\lambda}(\cdot\mid s)}
\left[Q_c^i(s,h)\right].
\end{aligned}
\end{equation}
Suppose that $(\eta_i^\star,\lambda_i^\star)$ minimizes $g_i$ with
$\eta_i^\star>0$, and set
$q_i^\star=q_{\eta_i^\star,\lambda_i^\star}$. The temperature variable is
interior, whereas $\lambda$ is constrained to be nonnegative. First-order
optimality therefore yields
\begin{equation}
\begin{aligned}
\mathbb{E}_{s\sim d_i}
D_{\mathrm{KL}}\!\left(
q_i^\star(\cdot\mid s)\,\|\,\pi_i(\cdot\mid s)
\right)
&=\epsilon_{\mathrm{KL}},
\\
\mathbb{E}_{s\sim d_i,\,h\sim q_i^\star(\cdot\mid s)}
\left[Q_c^i(s,h)\right]
&\le\epsilon_{E,i},
\\
\lambda_i^\star
\left(
\mathbb{E}_{s\sim d_i,\,h\sim q_i^\star(\cdot\mid s)}
\left[Q_c^i(s,h)\right]
-\epsilon_{E,i}
\right)
&=0.
\end{aligned}
\end{equation}
These are primal feasibility and complementary slackness. Since
$q_i^\star$ maximizes the Lagrangian at the dual optimum, zero duality gap
establishes its primal optimality and completes the proof of
Theorem~\ref{thm:hybrid_teacher}.

\subsection{Derivation of Hybrid-Action KL Decomposition}
\label{app:proof_hybrid_kl}

Consider the joint densities $q,\pi$ and their factors
$q_m,\pi_m,q_c,\pi_c$ introduced in Section~\ref{sec:hybrid_kl}. The mode
set $\mathcal M$, conditional action spaces $\mathcal A_c^m$, dimensions
$p_m$, and reference-measure convention are as defined in
Equations~(\ref{eq:hybrid_action_measure})--
(\ref{eq:hybrid_reference_measure}). Fix a state $s$ such that
$q(\cdot\mid s)\ll\pi(\cdot\mid s)$. This absolute continuity implies
$q_m(\cdot\mid s)\ll\pi_m(\cdot\mid s)$ and, for every mode satisfying
$q_m(m\mid s)>0$,
$q_c(\cdot\mid s,m)\ll\pi_c(\cdot\mid s,m)$. Hence, for
$q(\cdot\mid s)$-a.e. $(m,a_c)$,
\begin{equation}
\log\frac{q(m,a_c\mid s)}{\pi(m,a_c\mid s)}
=
\log\frac{q_m(m\mid s)}{\pi_m(m\mid s)}
+
\log\frac{q_c(a_c\mid s,m)}{\pi_c(a_c\mid s,m)}.
\end{equation}
Integrating this identity under $q(\cdot\mid s)$ and using
$\int_{\mathcal A_c^m}q_c(a_c\mid s,m)\,da_c=1$ gives
\begin{equation}
\begin{aligned}
&D_{\mathrm{KL}}\!\left(
q(\cdot\mid s)\,\|\,\pi(\cdot\mid s)
\right)
\\
={}&
\sum_{m\in\mathcal M}
q_m(m\mid s)
\int_{\mathcal A_c^m}
q_c(a_c\mid s,m)
\left[
\log\frac{q_m(m\mid s)}{\pi_m(m\mid s)}
+
\log\frac{q_c(a_c\mid s,m)}{\pi_c(a_c\mid s,m)}
\right]
\,da_c
\\
={}&
\sum_{m\in\mathcal M}
q_m(m\mid s)
\log\frac{q_m(m\mid s)}{\pi_m(m\mid s)}
\\
&+
\sum_{m\in\mathcal M}
q_m(m\mid s)
D_{\mathrm{KL}}\!\left(
q_c(\cdot\mid s,m)\,\|\,\pi_c(\cdot\mid s,m)
\right).
\end{aligned}
\end{equation}
The first sum is
$D_{\mathrm{KL}}(q_m(\cdot\mid s)\,\|\,\pi_m(\cdot\mid s))$, and the second is
the $q_m(\cdot\mid s)$-expectation of the mode-conditioned KL divergence.
Terms with $q_m(m\mid s)=0$ are assigned value zero. The identity also
holds when a KL divergence is infinite: the negative parts of the
log-density ratios are integrable, since $-t\log t\le1/e$ for
$0<t\le1$, so the decomposition involves no subtraction of infinities.
This proves Equation~(\ref{eq:hybrid_kl_decomposition}); the unit-mass convention also
covers modes with $p_m=0$.

\subsection{Derivation of Factorized Variational Projection}
\label{app:proof_projection}

As in Section~\ref{sec:hybrid_kl}, assume finite log-density expectations
under $d_i(ds)q_i^\star(dh\mid s)$. By
Equation~(\ref{eq:hybrid_policy_pre}), the hybrid policy satisfies the
following identity almost everywhere under this measure:
\begin{equation}
\log\pi_\theta(m,a_c\mid s)
=
\log\pi_{\theta_m}(m\mid s)
+
\log\pi_{\theta_c}(a_c\mid s,m).
\end{equation}
Recall that $q_{i,m}^\star$ is the mode marginal of the joint target
$q_i^\star$ in Equation~(\ref{eq:hcvpo_teacher}); for $d_i$-a.e. $s$,
\begin{equation}
q_{i,m}^\star(m\mid s)
=
\int_{\mathcal A_c^m}q_i^\star(m,a_c\mid s)\,da_c.
\end{equation}
Since $\log\pi_{\theta_m}(m\mid s)$ is independent of $a_c$, linearity of
expectation and the preceding marginalization give
\begin{equation}
\begin{aligned}
&\mathbb{E}_{s\sim d_i,\,(m,a_c)\sim q_i^\star(\cdot\mid s)}
\left[
\log\pi_\theta(m,a_c\mid s)
\right]
\\
={}&
\mathbb{E}_{s\sim d_i,\,m\sim q_{i,m}^\star(\cdot\mid s)}
\left[
\log\pi_{\theta_m}(m\mid s)
\right]
\\
&+
\mathbb{E}_{s\sim d_i,\,(m,a_c)\sim q_i^\star(\cdot\mid s)}
\left[
\log\pi_{\theta_c}(a_c\mid s,m)
\right].
\end{aligned}
\end{equation}
The discrete-policy term therefore depends only on $q_{i,m}^\star$, whereas
the conditional-continuous term retains the full joint target
$q_i^\star$. This is Equation~(\ref{eq:factorized_projection}).

\subsection{Existence of an Approximate Dual Minimizer}
\label{app:proof_bounded_dual}

By Equation~(\ref{eq:bounded_dual_domain}), $\mathcal D_i$ is the Cartesian
product of the nonempty closed intervals
$[\eta_{\min},\eta_{\max}]$ and $[0,\lambda_{\max}]$. It is therefore closed
and bounded in $\mathbb R^2$, and hence compact by the Heine--Borel theorem.
The approximate dual objective
$\widehat g_i:\mathcal D_i\to\mathbb R$ is assumed continuous in
Section~\ref{sec:dual_stabilization}. The extreme-value theorem then yields
a pair $(\widehat\eta_i,\widehat\lambda_i)\in\mathcal D_i$ satisfying
\begin{equation}
\widehat g_i(\widehat\eta_i,\widehat\lambda_i)
=
\min_{(\eta,\lambda)\in\mathcal D_i}
\widehat g_i(\eta,\lambda).
\end{equation}
Thus, the bounded approximate dual admits a minimizer, as claimed in
Section~\ref{sec:dual_stabilization}.

\subsection{Proof of Fixed-Reference E/M Surrogate Improvement}
\label{app:proof_fixed_reference}

We use the exact-update and finiteness assumptions of
Proposition~\ref{prop:fixed_reference_improvement}, including the finite
M-step log-density expectations specified in Section~\ref{sec:hybrid_kl}.
All quantities defining the fixed-reference surrogate---$d_i$, $Q_r^i$,
$Q_c^i$, the KL multiplier $\eta_i^\star>0$, and the cost multiplier
$\lambda_i^\star\ge0$---are held fixed throughout the E/M cycle.
Let $q$ be any stochastic kernel from $\mathcal S$ to $\mathcal H$ satisfying
$q(\cdot\mid s)\ll\pi_i(\cdot\mid s)$ for $d_i$-a.e. $s$ and having finite
$\mathcal F_i(q,\pi_i)$. Equations~(\ref{eq:hcvpo_lagrangian}) and
(\ref{eq:fixed_reference_surrogate}) give
\begin{equation}
\mathcal L_i(q,\eta_i^\star,\lambda_i^\star)
=
\mathcal F_i(q,\pi_i)
+\lambda_i^\star\epsilon_{E,i}
+\eta_i^\star\epsilon_{\mathrm{KL}}.
\end{equation}
The last two terms are independent of $q$, so the Lagrangian and
$\mathcal F_i(\cdot,\pi_i)$ have the same maximizers on this domain. By
Theorem~\ref{thm:hybrid_teacher}, the exact E-step target $q_i^\star$ is such
a maximizer. Therefore,
\begin{equation}
\mathcal F_i(q_i^\star,\pi_i)
\ge
\mathcal F_i(q,\pi_i)
\end{equation}
for every such $q$.

Next fix $q_i^\star$. The current policy $\pi_i$ is M-step feasible because
both old-to-new component KL divergences in Equation~(\ref{eq:hcvpo_mstep})
vanish when evaluated at $\pi_i$. Hence the exact M-step solution
$\pi_{i+1}$ satisfies
\begin{equation}
\mathbb{E}_{s\sim d_i,\,h\sim q_i^\star(\cdot\mid s)}
\left[
\log\pi_{i+1}(h\mid s)-\log\pi_i(h\mid s)
\right]
\ge 0.
\end{equation}
For fixed $q_i^\star$, only the KL term in
$\mathcal F_i(q_i^\star,\pi)$ depends on $\pi$. Expanding its difference
gives
\begin{equation}
\begin{aligned}
&\mathcal F_i(q_i^\star,\pi_{i+1})
-\mathcal F_i(q_i^\star,\pi_i)
\\
={}&
\eta_i^\star
\mathbb{E}_{s\sim d_i,\,h\sim q_i^\star(\cdot\mid s)}
\left[
\log\pi_{i+1}(h\mid s)-\log\pi_i(h\mid s)
\right]
\ge 0.
\end{aligned}
\end{equation}
Combining the two inequalities gives the chain in
Proposition~\ref{prop:fixed_reference_improvement}.

\subsection{Proof of Conditional Reward and Cost Bounds}
\label{app:proof_cost_bound}

We use the occupancies $d_{\pi_i},d_{\pi_{i+1}}$, returns $J_x$,
advantages $A_x^{\pi_i}$, and statewise policy shift $\Delta_i$ from
Section~\ref{sec:theory_properties}, with $x\in\{r,c\}$ indexing reward
and cost. These are true normalized discounted occupancies, distinct from
the fixed reference $d_i$. The transition kernel $P$, initial-state
distribution $\rho_0$, discount factor $\gamma$, and cost limit $d$ are
as defined in Section~\ref{sec:preliminaries}. Define the new-policy
expected advantage relative to $\pi_i$ by
\begin{equation}
\bar A_{x,i+1}(s)
=
\mathbb{E}_{h\sim\pi_{i+1}(\cdot\mid s)}
\left[A_x^{\pi_i}(s,h)\right].
\end{equation}
By the definition of $B_{x,i+1}$ in Section~\ref{sec:theory_properties},
$|\bar A_{x,i+1}(s)|\le B_{x,i+1}$ for every $s\in\mathcal S$.
The performance-difference identity, followed by adding and subtracting the
old-policy occupancy term, gives
\begin{equation}
\begin{aligned}
J_x(\pi_{i+1})-J_x(\pi_i)
={}&
\frac{1}{1-\gamma}
\mathbb{E}_{s\sim d_{\pi_{i+1}}}
\left[\bar A_{x,i+1}(s)\right]
\\
={}&
\frac{1}{1-\gamma}
\mathbb{E}_{s\sim d_{\pi_i}}
\left[\bar A_{x,i+1}(s)\right]
+\mathcal R_{x,i},
\end{aligned}
\end{equation}
where the occupancy correction is
\begin{equation}
\mathcal R_{x,i}
=
\frac{1}{1-\gamma}
\int_{\mathcal S}
\bar A_{x,i+1}(s)
\left(d_{\pi_{i+1}}-d_{\pi_i}\right)(ds).
\end{equation}
For a finite signed measure $\sigma$ on $\mathcal S$, let $|\sigma|$ denote
its variation measure and define its total-variation norm by
$\|\sigma\|_1=|\sigma|(\mathcal S)$. The preceding pointwise bound gives
\begin{equation}
|\mathcal R_{x,i}|
\le
\frac{B_{x,i+1}}{1-\gamma}
\left\|d_{\pi_{i+1}}-d_{\pi_i}\right\|_1.
\end{equation}

It remains to control this occupancy shift. For a policy $\pi$ and a
measurable state set $E\subseteq\mathcal S$, define the induced state
transition kernel by
\begin{equation}
P_\pi(E\mid s)
=
\int_{\mathcal H}P(E\mid s,h)\pi(dh\mid s).
\end{equation}
For such a signed measure $\sigma$, define its propagation under $P_\pi$ by
\begin{equation}
\left(P_\pi^\top\sigma\right)(E)
=
\int_{\mathcal S}P_\pi(E\mid s)\,\sigma(ds).
\end{equation}
Thus, $P_\pi^\top$ is the measure-propagation operator associated with the
policy-induced transition kernel. As a Markov operator, it contracts total
variation:
\begin{equation}
\left\|P_\pi^\top\sigma\right\|_1
\le
\|\sigma\|_1.
\end{equation}
Moreover, the data-processing inequality for total variation gives, for
every $s\in\mathcal S$,
\begin{equation}
\begin{aligned}
\left\|
P_{\pi_{i+1}}(\cdot\mid s)-P_{\pi_i}(\cdot\mid s)
\right\|_1
&=
2D_{\mathrm{TV}}\!\left(
P_{\pi_{i+1}}(\cdot\mid s),P_{\pi_i}(\cdot\mid s)
\right)
\\
&\le
2D_{\mathrm{TV}}\!\left(
\pi_{i+1}(\cdot\mid s),\pi_i(\cdot\mid s)
\right)
\le 2\Delta_i .
\end{aligned}
\end{equation}
Because $d_{\pi_i}$ is a probability measure, this statewise bound implies
\begin{equation}
\begin{aligned}
&\left\|
\left(P_{\pi_{i+1}}^\top-P_{\pi_i}^\top\right)d_{\pi_i}
\right\|_1
\\
&\qquad\le
\int_{\mathcal S}
\left\|
P_{\pi_{i+1}}(\cdot\mid s)-P_{\pi_i}(\cdot\mid s)
\right\|_1
d_{\pi_i}(ds)
\le 2\Delta_i.
\end{aligned}
\end{equation}
The normalized discounted occupancy obeys the fixed-point equation
\begin{equation}
d_\pi
=
(1-\gamma)\rho_0+\gamma P_\pi^\top d_\pi.
\end{equation}
Subtracting the equations for $\pi_{i+1}$ and $\pi_i$ gives
\begin{equation}
\begin{aligned}
d_{\pi_{i+1}}-d_{\pi_i}
={}&
\gamma P_{\pi_{i+1}}^\top
\left(d_{\pi_{i+1}}-d_{\pi_i}\right)
\\
&+
\gamma
\left(P_{\pi_{i+1}}^\top-P_{\pi_i}^\top\right)d_{\pi_i}.
\end{aligned}
\end{equation}
Applying the two preceding bounds yields
\begin{equation}
\begin{aligned}
\left\|d_{\pi_{i+1}}-d_{\pi_i}\right\|_1
\le{}&
\gamma
\left\|d_{\pi_{i+1}}-d_{\pi_i}\right\|_1
\\
&+
\gamma
\left\|
\left(P_{\pi_{i+1}}^\top-P_{\pi_i}^\top\right)d_{\pi_i}
\right\|_1
\\
\le{}&
\gamma
\left\|d_{\pi_{i+1}}-d_{\pi_i}\right\|_1
+2\gamma\Delta_i .
\end{aligned}
\end{equation}
Rearranging gives
\begin{equation}
\left\|d_{\pi_{i+1}}-d_{\pi_i}\right\|_1
\le
\frac{2\gamma}{1-\gamma}\Delta_i.
\end{equation}
Substitution into the occupancy-correction bound yields
\begin{equation}
|\mathcal R_{x,i}|
\le
\frac{2\gamma B_{x,i+1}}{(1-\gamma)^2}\Delta_i .
\end{equation}

Finally, the definition of $\bar A_{x,i+1}$ gives
\begin{equation}
\mathbb{E}_{s\sim d_{\pi_i}}
\left[\bar A_{x,i+1}(s)\right]
=
\mathbb{E}_{\substack{s\sim d_{\pi_i}\\
h\sim\pi_{i+1}(\cdot\mid s)}}
\left[A_x^{\pi_i}(s,h)\right].
\end{equation}
Using
$\mathcal R_{r,i}\ge
-2\gamma B_{r,i+1}\Delta_i/(1-\gamma)^2$ for reward and
$\mathcal R_{c,i}\le
2\gamma B_{c,i+1}\Delta_i/(1-\gamma)^2$ for cost therefore proves both
stated inequalities.

With $U_{c,i}$ denoting the cost-increase upper bound defined in
Section~\ref{sec:theory_properties}, Proposition~\ref{prop:cost_variation}
gives $J_c(\pi_{i+1})\le J_c(\pi_i)+U_{c,i}$. If
$J_c(\pi_i)\le d-\zeta$ for a margin $\zeta>0$ and $U_{c,i}\le\zeta$, then
\begin{equation}
J_c(\pi_{i+1})
\le
J_c(\pi_i)+U_{c,i}
\le
(d-\zeta)+\zeta
=d.
\end{equation}
This proves the sufficient feasibility-margin statement following
Proposition~\ref{prop:cost_variation}.

\section{Training and Experimental Details}
\label{app:experimental_details}

This section details the implementation of H-CVPO developed in
Section~\ref{sec:hcvpo} and summarized in Algorithm~\ref{alg:hcvpo_training}.
It also specifies the tasks and shared training and evaluation settings
used in Section~\ref{sec:experiments}.

\subsection{H-CVPO Training Procedure}
\label{app:hcvpo_training_protocol}

H-CVPO combines off-policy critic learning with the fixed-reference E/M
update in Section~\ref{sec:hcvpo}. Algorithm~\ref{alg:hcvpo_training}
summarizes transition collection, critic and target-critic updates, and
scheduled variational updates after the prescribed critic warm-up.

\paragraph{Fixed-reference cycle.}
Each E/M cycle uses a uniform reference batch $\{s_b\}_{b=1}^{B}$ drawn from
replay, separately from the samples used for critic regression. This batch
provides the empirical counterpart of $d_i$: the same batch averages are
used in the E-step, M-step, and empirical KL checks, regardless of the
critic sampling scheme. At \emph{Freeze}, the current actor is copied to
$\pi_i$, and the reference batch, $\pi_i$, and critic estimates $Q_r^i,Q_c^i$
are held fixed throughout the cycle. The sampled particles remain fixed
throughout the probe and any subsequent dual solve; the resulting target
$\widehat q_i$ remains fixed during actor fitting. Thus, actor parameters may
change in the M-step, but the reference policy does not. Throughout this
subsection, $Q_c^i$ denotes the frozen cost critic and plays the role of
$\widehat Q_c^i$ in Equation~(\ref{eq:approximate_residuals}).

\paragraph{E-step: joint target construction.}
For each reference state $s_b$ and mode $m\in\mathcal M$, draw a positive
integer number $K_{b,m}$ of continuous particles from the old conditional
policy:
\begin{equation}
a_{b,m,j}
\sim
\pi_{i,c}(\cdot\mid s_b,m),
\qquad
j=1,\ldots,K_{b,m}.
\label{eq:practical_particle_sampling}
\end{equation}
The corresponding hybrid candidate
$h_{b,m,j}=(m,a_{b,m,j})$ is assigned the old-policy quadrature mass
\begin{equation}
\omega_{b,m,j}^{i}
=
\frac{\pi_{i,m}(m\mid s_b)}{K_{b,m}}.
\label{eq:practical_particle_mass}
\end{equation}
The masses for a given mode sum to its old-policy probability, preserving
the discrete marginal rather than replacing it with a uniform distribution.
The \emph{Draw} and \emph{Form} steps therefore provide a quadrature rule for
the hybrid sum--integral in Equation~(\ref{eq:hcvpo_partition_general}).
The \emph{Evaluate} step applies the frozen critics to every candidate.

The reward-oriented probe sets $\lambda=0$; let
$\widehat C_i^{(0)}$ denote the resulting batch-weighted cost used to test the
E-step constraint. The algorithm uses the probe solution with
$\widehat\lambda_i=0$ when
$\widehat C_i^{(0)}\le\rho\epsilon_{E,i}$, where $\rho\in(0,1)$ is a
dead-zone fraction that places the probe acceptance threshold below
$\epsilon_{E,i}$.
Otherwise, the algorithm solves the bounded approximate dual over
$\mathcal D_i$, yielding
$(\widehat\eta_i,\widehat\lambda_i)\in\mathcal D_i$ for the reference batch.
This bounded solve approximates Equation~(\ref{eq:hcvpo_dual}) on the domain
in Equation~(\ref{eq:bounded_dual_domain}). In either branch,
$\widehat\eta_i$ controls the sharpness of reweighting; a positive
$\widehat\lambda_i$ discounts costly candidates. The surrogate threshold
$\epsilon_{E,i}$ is distinct from the episodic CMDP limit $d$. When enabled,
the prior-support safeguard is applied only in the inactive-cost branch
($\widehat\lambda_i=0$), before target construction. For each reference state
$s_b$, the particle weights are normalized as
\begin{equation}
\widehat w_{b,m,j}^{i}
=
\frac{
\omega_{b,m,j}^{i}
\exp\!\left(
\frac{
Q_r^i(s_b,h_{b,m,j})
-\widehat\lambda_iQ_c^i(s_b,h_{b,m,j})
}{
\widehat\eta_i
}
\right)
}{
\displaystyle
\sum_{m'\in\mathcal M}\sum_{j'=1}^{K_{b,m'}}
\omega_{b,m',j'}^{i}
\exp\!\left(
\frac{
Q_r^i(s_b,h_{b,m',j'})
-\widehat\lambda_iQ_c^i(s_b,h_{b,m',j'})
}{
\widehat\eta_i
}
\right)
}.
\label{eq:practical_particle_weight}
\end{equation}
These weights sum to one over $(m,j)$ for each $b$. The
\emph{Construct $\widehat q_i$} step uses them to approximate expectations
under the joint target in Equation~(\ref{eq:hcvpo_teacher}). The particles
serve as quadrature points; the absolute-continuity requirement in
Section~\ref{sec:dual_stabilization} concerns the underlying target kernel,
not the atomic empirical measure formed by the sampled points.

\paragraph{M-step: target validation and projection.}
Before projection, the algorithm evaluates empirical counterparts of the
cost and target-to-old-policy KL residuals in
Equation~(\ref{eq:approximate_residuals}) on the fixed reference batch.
The target must satisfy the finiteness and support conditions in
Section~\ref{sec:dual_stabilization}, with residuals no larger than
$\tau_c$ and $\tau_{\mathrm{KL}}$, respectively. If this admissibility test
fails, the \emph{Skip} step leaves the actor unchanged for the cycle.
Selecting the reward-oriented probe does not bypass this test.

For an admissible target, the algorithm saves the current actor parameters
and performs $M_{\mathrm M}$ updates using the empirical weighted
log-likelihood in Equation~(\ref{eq:factorized_projection}). Both actor
factors are fitted to the same fixed joint target. The \emph{Measure} step
then evaluates the old-to-new KL divergences in
Equation~(\ref{eq:hcvpo_mstep}) on the reference batch. The discrete term
compares mode probabilities; the conditional-continuous term averages the
mode-conditioned KL divergences under the old mode probabilities
$\pi_{i,m}$, as required by Equation~(\ref{eq:hybrid_kl_decomposition}).
The candidate is accepted only when all measured quantities are finite and
both trust-region limits hold; otherwise, the saved actor parameters are
restored. These post-update checks are distinct from the pre-projection
target-admissibility test. The continuous mean and standard-deviation
controls used in the experiments are reported in
Table~\ref{tab:hcvpo_hyperparameters}. The algorithm returns the policy at
the predefined training endpoint, consistent with the checkpoint protocol
in Section~\ref{sec:tasks_baselines}.

\begin{algorithm}[t]
\caption{Practical H-CVPO training procedure.}
\label{alg:hcvpo_training}
\small
\begin{algorithmic}[1]
\Require Hybrid actor $\pi_\theta$ with factors
$\pi_{\theta_m},\pi_{\theta_c}$;
reward and cost critics and their target copies;
replay buffer $\mathcal B_{\mathrm{rep}}$;
reference batch size $B$;
E-step limits $\epsilon_{E,i},\epsilon_{\mathrm{KL}}$;
admissibility tolerances $\tau_c,\tau_{\mathrm{KL}}$;
M-step limits $\epsilon_m,\epsilon_c$;
dual domain $\mathcal D_i$;
dead-zone fraction $\rho\in(0,1)$;
particle counts $\{K_{b,m}\}$;
M-step update count $M_{\mathrm M}$
\Ensure Hybrid policy $\pi_\theta$ at the predefined final training step

\State Initialize the actor, critics, target critics, replay buffer, and
optimization states

\While{the training interaction budget is not exhausted}

    \State Collect complete transitions
    $(s_t,h_t,r_t,c_t,s_{t+1})$ and termination flags using $\pi_\theta$
    \State Append the transitions to $\mathcal B_{\mathrm{rep}}$
    \State Update the reward and cost critics from replay
    \State Update the target critics by Polyak averaging

    \If{a variational update is scheduled and the prescribed critic warm-up is complete}

        \State Sample a uniform reference batch
        $\{s_b\}_{b=1}^{B}$ from $\mathcal B_{\mathrm{rep}}$
        \State \emph{Freeze}
        $\pi_i\leftarrow\pi_\theta$, the reference batch, and
        $Q_r^i,Q_c^i$ for this E/M cycle

        \Statex \textbf{E-step: construct the joint hybrid target}

        \For{$b=1,\ldots,B$ and each $m\in\mathcal M$}
            \State \emph{Draw} the particles in
            Equation~(\ref{eq:practical_particle_sampling})
            \State \emph{Form} $h_{b,m,j}=(m,a_{b,m,j})$ and assign the masses in
            Equation~(\ref{eq:practical_particle_mass})
            \State \emph{Evaluate}
            $Q_r^i(s_b,h_{b,m,j})$ and
            $Q_c^i(s_b,h_{b,m,j})$
        \EndFor

        \State Solve the reward-oriented probe with $\lambda=0$ and compute
        its batch-weighted cost $\widehat C_i^{(0)}$
        \State If $\widehat C_i^{(0)}\le\rho\epsilon_{E,i}$, use the probe
        solution with $\widehat\lambda_i=0$; otherwise solve the bounded
        approximate dual over $\mathcal D_i$
        \State When enabled, apply a prior-support safeguard within the
        inactive-cost branch ($\widehat\lambda_i=0$)
        \State \emph{Construct $\widehat q_i$} using the weights in
        Equation~(\ref{eq:practical_particle_weight})

        \Statex \textbf{M-step: validate and project the joint target}

        \If{$\widehat q_i\notin\mathcal Q_{i,\mathrm{valid}}$}
            \State \emph{Skip} the projection and leave the actor unchanged
        \Else
            \State Save the current actor parameters
            \State Perform $M_{\mathrm M}$ updates of both actor factors by
            weighted joint maximum likelihood on the fixed target and reference batch
            \State \emph{Measure} the empirical discrete and conditional-continuous
            old-to-new KL divergences
            \If{all measured quantities are finite and both trust-region
            limits hold}
                \State \emph{Accept} the candidate
            \Else
                \State Restore the saved actor parameters
            \EndIf
        \EndIf

    \EndIf

\EndWhile

\State \Return the policy at the predefined final training step
\end{algorithmic}
\end{algorithm}

\subsection{Experimental Details}
\label{app:implementation_details}
\label{app:tasks}

\paragraph{Comparison protocol.}
\label{app:comparison_protocol}
We compare all methods empirically and separately discuss their theoretical
guarantees. Under the stated assumptions, Theorem~\ref{thm:hybrid_teacher}
establishes E-step strong duality for H-CVPO, while
Section~\ref{sec:theory_properties} provides fixed-reference surrogate
improvement and conditional reward--cost bounds for its exact E/M updates.
CHPO switches between reward and cost
optimization; its analysis derives finite-iteration bounds on the
policy-objective gap and the excess of expected cost over its threshold
under its own assumptions~\citep{Zhou2025CHPO}. We make no additional
convergence or feasibility claims for HPPO-Lag or PDQN-Rco.

For the standard $320$k-step experiments, every final-policy table evaluates
the policy obtained at that fixed endpoint for each method and training
seed, following Section~\ref{sec:tasks_baselines}. We do not substitute an
earlier checkpoint on the basis of its reward or estimated feasibility.
This keeps the interaction budget and checkpoint-selection rule matched
across methods.

Following the task setup of CHPO~\citep{Zhou2025CHPO}, we use constrained
versions of
\emph{Moving}, \emph{Sliding}, and \emph{HardMove}, three hybrid-action
benchmarks from DI-engine, together with the parameterized \emph{Parking}
task considered in that work. Each task pairs a discrete mode with a
continuous control valid for that mode, consistent with
Equation~(\ref{eq:hybrid_action_measure}). The tasks differ in observation
and action dimensions, dynamics, and cost definitions.
Table~\ref{tab:task_protocol} summarizes their interfaces, effective episode
horizons, and primary experimental settings. The primary cost limits are
denoted by $d$, as in Equation~(\ref{eq:cmdp_pre}).

\begin{table}[H]
\centering
\scriptsize
\setlength{\tabcolsep}{1.2pt}
\renewcommand{\arraystretch}{1.16}
\caption{Task interfaces and primary experimental settings.}
\label{tab:task_protocol}
\begin{tabular}{
|C{0.11\columnwidth}
|C{0.11\columnwidth}
|C{0.08\columnwidth}
|C{0.25\columnwidth}
|C{0.11\columnwidth}
|C{0.09\columnwidth}
|C{0.14\columnwidth}|
}
\hline
\textbf{Task}
& \textbf{Observation dimension}
& \textbf{Modes}
& \textbf{Continuous action}
& \textbf{Episode horizon}
& \textbf{Primary $d$}
& \textbf{Training steps} \\
\hline
Moving
& $18$
& $3$
& 2-D, $[0,1]\times[-1,1]$
& $201$
& $1$
& $320{,}000$ \\
\hline
Sliding
& $20$
& $3$
& 2-D, $[0,1]\times[-1,1]$
& $201$
& $1$
& $320{,}000$ \\
\hline
HardMove
& $18$
& $16$
& 4-D, $[-1,1]^4$, mode-masked
& $26$
& $1$
& $320{,}000$ \\
\hline
Parking
& $17$
& $3$
& 1-D, $[-1,1]$
& $50$
& $1$
& $320{,}000$ \\
\hline
\end{tabular}
\end{table}

We next describe the task-specific dynamics and reward--cost definitions used in our experiments.

\paragraph{Moving.}
The agent must reach and stop within a target region while avoiding circular
hazards. Each action selects acceleration, turning, or braking and includes
$a_c\in[0,1]\times[-1,1]$. Acceleration uses the first continuous component,
turning uses the second, and braking uses neither; inactive components do
not affect actuation. Let $\delta_t$ denote the distance to the target.
On ordinary transitions, the reward $r_t=\delta_{t-1}-\delta_t$ measures
progress toward the target. The per-step cost equals twice the number of
circular hazard regions
occupied by the agent. Success requires the agent to be within the target
radius with zero speed. Episodes end on success, departure from the field,
or expiry of the time limit.

\paragraph{Sliding.}
Sliding retains Moving's objective, three modes, and two-dimensional
continuous control, but adds inertia: motion combines the vector induced by
the current action with the previous velocity vector. The experimental
observation has $20$ dimensions, augmenting the underlying $18$-dimensional
observation with $(v_x,v_y)=(v\cos\phi,v\sin\phi)$, where $v$ and $\phi$
denote speed and heading. Rewards and success conditions are unchanged
from Moving, as are the out-of-bounds and time-limit conditions for ending
an episode. The per-step cost, however, equals the number of circular
hazard regions occupied by the agent, without Moving's factor of two.

\paragraph{HardMove.}
HardMove expands the discrete mode space through four binary actuators.
Its $16$ modes enumerate all on--off masks, paired with continuous control
$a_c\in[-1,1]^4$. The selected mask determines which continuous components
are active; the others are set to zero and do not affect actuation. Motion
is the sum of the displacement vectors generated by the active actuators.
The agent must reach the target while avoiding circular hazards. Reward
is based on distance reduction, with a success bonus and a negative terminal
reward for failure or timeout. The per-step cost is ten times the number
of circular hazard regions occupied by the agent. Reaching the target
constitutes success; episodes end on success, departure from the field, or
expiry of the time limit.

\paragraph{Parking.}
Parking requires perpendicular parking within prescribed position and
heading tolerances. Its three modes are right, straight, and left, with
$a_c\in[-1,1]$ specifying travel under the selected maneuver. Right and left
produce radius-$5$ arc motion through a kinematic center update; straight
produces linear travel. Let $d_{\mathrm{tar}}$ denote the distance to the
target, $\Delta\psi$ the heading error, and $r_{\mathrm{target}}$ the
successful-parking indicator. The per-step reward is
\begin{equation}
\label{eq:parking_reward}
r_t
=
0.5\left(-\frac{d_{\mathrm{tar}}}{\sqrt{10^2+17.6^2}}\right)
-\frac{|\Delta\psi|}{\pi}
+4r_{\mathrm{target}}.
\end{equation}
Collision, overlap, and timeout penalties have zero weight in the reward.
The per-step cost combines collision and proximity penalties with an overall
scale factor of $1/10$; the proximity penalty activates when the obstacle
distance is below $0.2$. Let $\Delta x$ and $\Delta y$ denote target-relative
position errors. Successful parking requires $|\Delta x|<0.15$,
$|\Delta y|<0.35$, and $|\Delta\psi|<10^\circ$. Success terminates the
episode, whereas collision or proximity cost alone does not. Episodes
that do not end in success are truncated after $50$ steps.

\paragraph{Training configuration.}
All experiments are run on CPU using 32 vCPUs of an AMD EPYC 9654 processor
and 60 GB of memory. Table~\ref{tab:hcvpo_hyperparameters} reports the
H-CVPO network architecture and selected training settings for
Algorithm~\ref{alg:hcvpo_training}. The mode-policy M-step limit is
$\epsilon_m$. Separate limits on the continuous mean and standard deviation
implement the conditional-continuous trust-region checks represented
abstractly by $\epsilon_c$ in Equation~(\ref{eq:hcvpo_mstep}).

\begin{table}[H]
\centering
\scriptsize
\setlength{\tabcolsep}{3pt}
\renewcommand{\arraystretch}{1.08}
\caption{H-CVPO network architecture and selected training settings.}
\label{tab:hcvpo_hyperparameters}
\begin{tabular}{
|C{0.36\columnwidth}
|C{0.54\columnwidth}|
}
\hline
\textbf{Setting}
& \textbf{Value} \\
\hline
Policy network
& $[128,128]$, ReLU \\
\hline
Reward critic network
& $[256,256]$, ReLU \\
\hline
Cost critic network
& $[256,256]$, ReLU \\
\hline
Optimizer
& Adam \\
\hline
Learning rates
& actor $3\times10^{-4}$; reward and cost critics $10^{-3}$ \\
\hline
Critic / reference batch size
& $256/256$ \\
\hline
Replay capacity / critic warm-up
& $100{,}000$ transitions / $5{,}000$ environment steps \\
\hline
Discount factor
& $\gamma=0.99$ \\
\hline
Target-network update rate
& $0.005$ \\
\hline
E-step KL radius
& $\epsilon_{\mathrm{KL}}=0.02$ \\
\hline
M-step KL controls
& mode $\epsilon_m=0.01$; continuous mean $0.005$; continuous standard
deviation $0.0005$ \\
\hline
\end{tabular}
\end{table}

\subsection{Additional Baseline Comparisons}
\label{app:additional_comparisons}

We complement the main empirical comparison with three additional analyses.
First, we present complete learning trajectories under the cost limits
$d=1.5$ and $d=2.0$ to supplement the final-policy results.
Second, we compare the fixed-endpoint results with performance averaged over
the last five periodic evaluation checkpoints. These averages characterize
late-stage training behavior across successive policies; they do not
establish the feasibility of any single policy.
Third, we evaluate H-CVPO and CHPO over an extended training horizon while
varying the fraction of CHPO policy updates devoted to cost minimization,
as specified by the C/A ratio. This analysis assesses CHPO's sensitivity to
update allocation and whether the observed reward--cost differences between
the two methods persist under a larger interaction budget.

\subsubsection{Learning Performance under Additional Cost Limits}
\label{app:additional_cost_limits}

\begin{figure*}[t]
    \centering
    \includegraphics[width=\textwidth]
    {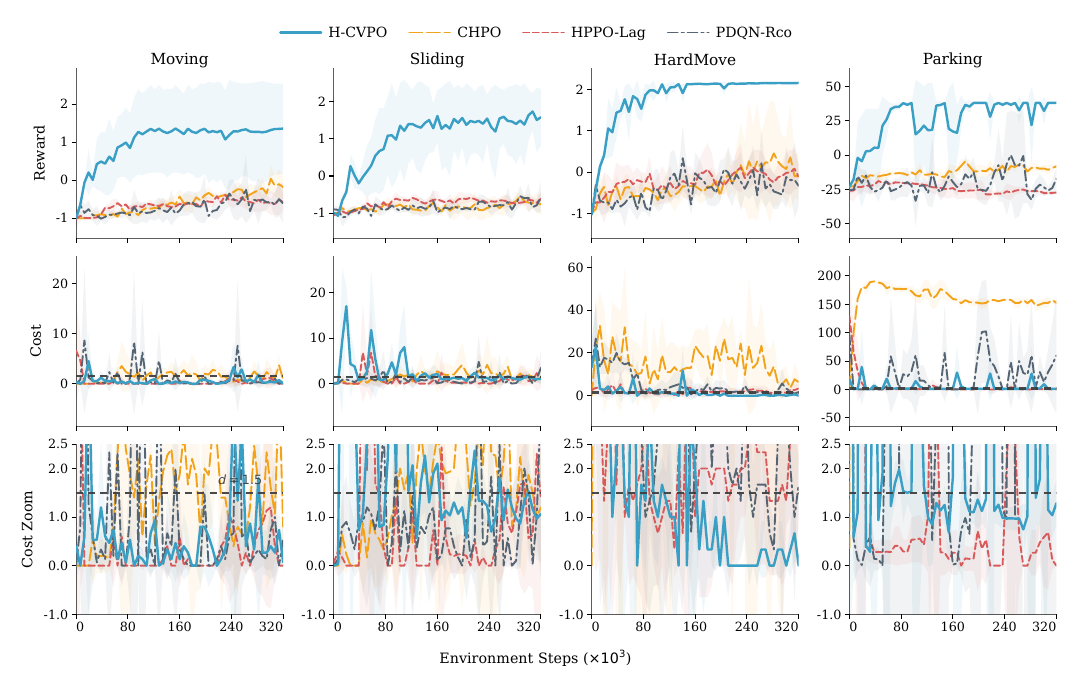}
    \caption{
    Learning curves for H-CVPO and the baselines under the cost limit
    $d=1.5$. Lines and shaded regions show the mean and one standard deviation
    over the three training seeds. Rows report reward, cost on the full scale,
    and cost near the constraint boundary; horizontal dashed lines mark the
    cost limit.
    }
    \label{fig:comparison_d15}
\end{figure*}

\begin{figure*}[t]
    \centering
    \includegraphics[width=\textwidth]
    {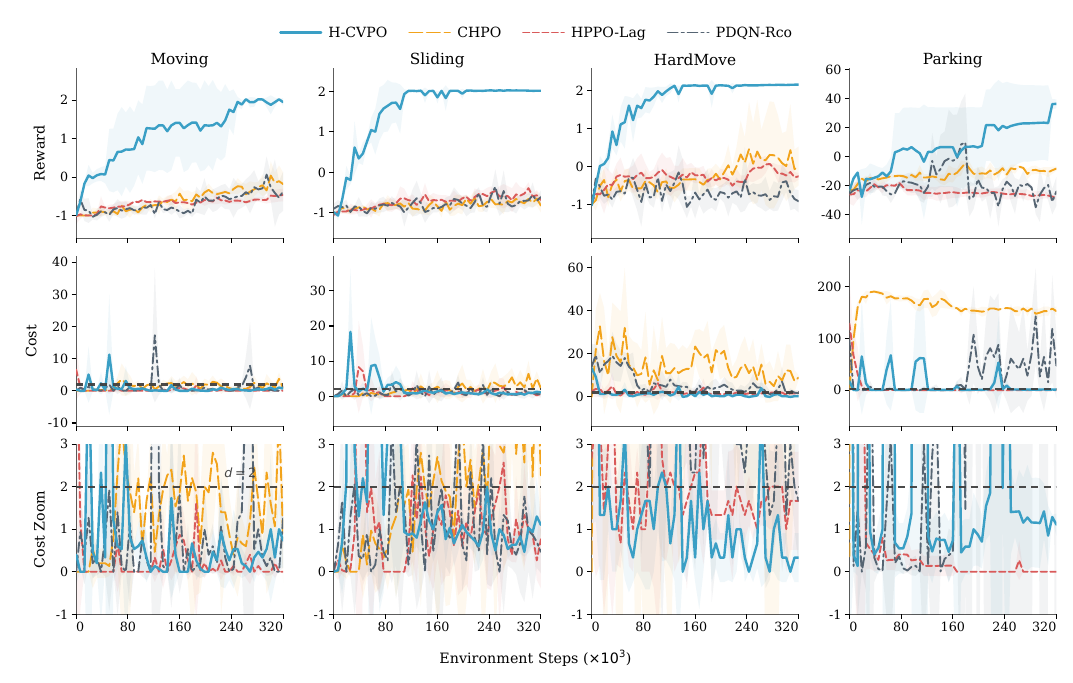}
    \caption{
    Learning curves for H-CVPO and the baselines under the cost limit
    $d=2.0$. Lines and shaded regions show the mean and one standard deviation
    over the three training seeds. Rows report reward, cost on the full scale,
    and cost near the constraint boundary; horizontal dashed lines mark the
    cost limit.
    }
    \label{fig:comparison_d20}
\end{figure*}

Figures~\ref{fig:comparison_d15} and~\ref{fig:comparison_d20} show the
reward and cost learning curves under the additional cost limits $d=1.5$
and $d=2.0$, complementing the final-policy results in
Table~\ref{tab:final40_main_neutral}. We analyze the reward and cost trends
shown in these two figures as follows.

\paragraph{Results at $d=1.5$.}
In Figure~\ref{fig:comparison_d15}, H-CVPO's mean reward settles near $1.3$
on Moving after about $100$k steps and fluctuates around $1.3$--$1.6$ on
Sliding. HardMove exceeds $2.0$ around $110$k steps and stays slightly above
that level. Parking reaches $35.0$--$38.0$ by $80$--$100$k steps and later
stays near $38.0$ with some fluctuations. Later mean costs are mostly below
$d=1.5$ on Moving and HardMove, although Moving has spikes near $240$k,
$260$k, and $280$k steps. HardMove's cost is near zero around
$210$--$260$k steps. Sliding's cost falls from early peaks but still sometimes
exceeds the limit. Parking's later cost is often $1.0$--$1.5$, with spikes near
$30.0$.

All baselines have substantially lower late-stage rewards. CHPO's costs
fluctuate above and below the limit on Moving and Sliding and remain above
it on HardMove and Parking, with Parking near $150.0$--$160.0$ later in training.
HPPO-Lag's later costs stay below the limit on Moving, sometimes exceed
it on Sliding and Parking, and frequently exceed it on HardMove.
PDQN-Rco's later Moving cost is usually below the limit, with a spike near
$250$k steps; Sliding's cost fluctuates above and below the limit, and
HardMove's often exceeds it. Parking has repeated large cost increases, including a peak
near $100.0$ around $200$--$210$k steps.

\paragraph{Results at $d=2.0$.}
In Figure~\ref{fig:comparison_d20}, H-CVPO's mean reward on Moving rises
from about $1.3$ around $110$k steps to $2.0$ around $230$--$250$k.
Sliding and HardMove reach about $2.0$ around $120$k steps, with HardMove
staying slightly higher. Parking reaches about $22.0$ around $210$k and
exceeds $35.0$ near the end. Mean costs on Moving and Sliding are at or
below $d=2.0$ throughout the second half of training; HardMove is mostly
below it, with occasional spikes. Parking has early cost spikes above
$50.0$ and another near $230$k, but after about $250$k its cost stays below
the limit, mostly around $1.0$--$1.5$.

All baselines have substantially lower late-stage rewards. CHPO's Moving
cost crosses $d=2.0$, while Sliding stays above it over roughly the last
$80$k steps. HardMove costs range from about $5.0$ to $15.0$ over that
interval, and Parking stays near $150.0$--$160.0$ in the second half.
HPPO-Lag's later costs are below the limit on Moving, mostly below it on
Sliding, between $1.0$ and $2.0$ on HardMove over the last $80$k, and near
zero on Parking. PDQN-Rco's later costs are mostly below the limit on
Moving and Sliding but often above it on HardMove. Parking's reward
increase is not sustained; costs repeatedly exceed $100.0$ and stay above
$2.0$ over roughly the last $80$k steps.

Overall, under both cost limits, H-CVPO achieves higher late-stage mean
rewards than the baselines across all four tasks. Its mean costs are often
below the corresponding limits later in training, although some tasks still
show cost spikes and exceed the limits at certain checkpoints. The baselines
have lower rewards without consistently better constraint control.

\subsubsection{Late-Stage Performance Trends}
\label{app:late_stage_robustness}

To assess whether the final checkpoint reflects nearby training behavior, we
average reward and cost over the five periodic evaluation checkpoints at
294.4k, 300.8k, 307.2k, 313.6k, and 320k environment steps for each training
seed, then report the mean and standard deviation across three seeds.
These averages combine different policies and describe late-stage performance,
not the feasibility of a single deployed policy.

\begin{table*}[t]
\centering
\scriptsize
\setlength{\tabcolsep}{2.4pt}
\caption{Late-stage performance over the final five periodic evaluation
checkpoints (294.4k--320k environment steps). For each training seed,
we average the five periodic evaluations, each computed from 10 deterministic
episodes with fixed evaluation seeds 21000--21009, and report the mean and
sample standard deviation across the three training seeds.}
\label{tab:late5_appendix_neutral}
\begin{tabular}{@{}ll*{8}{c}@{}}
\toprule
& & \multicolumn{2}{c}{H-CVPO} & \multicolumn{2}{c}{CHPO} & \multicolumn{2}{c}{HPPO-Lag} & \multicolumn{2}{c}{PDQN-Rco} \\
\cmidrule(lr){3-4}\cmidrule(lr){5-6}\cmidrule(lr){7-8}\cmidrule(l){9-10}
Environment & $d$ & Reward & Cost & Reward & Cost & Reward & Cost & Reward & Cost \\
\midrule
\multirow{3}{*}{Moving} & 1.0 & 2.02$\pm$0.01 & 0.69$\pm$0.29 & -0.14$\pm$0.19 & 1.03$\pm$0.85 & -0.67$\pm$0.05 & 0.37$\pm$0.65 & -0.62$\pm$0.44 & 0.48$\pm$0.49 \\
 & 1.5 & 1.33$\pm$1.22 & 0.35$\pm$0.10 & -0.15$\pm$0.18 & 1.91$\pm$1.24 & -0.58$\pm$0.30 & 0.55$\pm$0.95 & -0.58$\pm$0.20 & 0.48$\pm$0.25 \\
 & 2.0 & 1.95$\pm$0.05 & 0.72$\pm$0.84 & -0.15$\pm$0.18 & 1.91$\pm$1.24 & -0.50$\pm$0.14 & 0.04$\pm$0.07 & -0.30$\pm$0.45 & 0.35$\pm$0.18 \\
\midrule
\addlinespace[2pt]
\multirow{3}{*}{Sliding} & 1.0 & 1.20$\pm$0.57 & 3.21$\pm$4.35 & -0.62$\pm$0.17 & 2.42$\pm$3.47 & -0.68$\pm$0.14 & 0.66$\pm$0.94 & -0.88$\pm$0.13 & 0.55$\pm$0.30 \\
 & 1.5 & 1.57$\pm$0.79 & 1.21$\pm$0.55 & -0.72$\pm$0.21 & 1.38$\pm$1.00 & -0.67$\pm$0.10 & 1.31$\pm$2.14 & -0.67$\pm$0.14 & 1.59$\pm$0.46 \\
 & 2.0 & 2.02$\pm$0.00 & 0.97$\pm$0.37 & -0.73$\pm$0.13 & 3.71$\pm$1.53 & -0.55$\pm$0.01 & 0.85$\pm$0.43 & -0.66$\pm$0.11 & 0.95$\pm$0.36 \\
\midrule
\addlinespace[2pt]
\multirow{3}{*}{HardMove} & 1.0 & 2.14$\pm$0.00 & 0.60$\pm$0.60 & 0.01$\pm$0.85 & 10.40$\pm$4.33 & 0.24$\pm$0.86 & 2.53$\pm$1.70 & -0.78$\pm$0.21 & 1.73$\pm$1.15 \\
 & 1.5 & 2.15$\pm$0.01 & 0.27$\pm$0.23 & 0.07$\pm$0.77 & 6.13$\pm$4.71 & -0.03$\pm$0.40 & 2.33$\pm$2.14 & -0.28$\pm$0.37 & 2.73$\pm$2.50 \\
 & 2.0 & 2.15$\pm$0.01 & 0.27$\pm$0.46 & 0.09$\pm$0.78 & 9.67$\pm$6.52 & -0.23$\pm$0.18 & 1.60$\pm$1.04 & -0.66$\pm$0.30 & 3.00$\pm$1.91 \\
\midrule
\addlinespace[2pt]
\multirow{3}{*}{Parking} & 1.0 & 22.63$\pm$27.00 & 1.15$\pm$0.19 & -9.59$\pm$1.85 & 152.82$\pm$4.22 & -28.55$\pm$1.57 & 0.00$\pm$0.00 & -4.61$\pm$37.20 & 0.56$\pm$0.60 \\
 & 1.5 & 37.05$\pm$2.06 & 2.88$\pm$3.24 & -9.59$\pm$1.85 & 152.82$\pm$4.22 & -27.42$\pm$1.06 & 0.37$\pm$0.51 & -22.47$\pm$4.15 & 35.62$\pm$50.98 \\
 & 2.0 & 28.54$\pm$16.80 & 1.16$\pm$0.51 & -9.59$\pm$1.85 & 152.82$\pm$4.22 & -27.07$\pm$1.13 & 0.00$\pm$0.00 & -24.49$\pm$3.27 & 54.28$\pm$58.51 \\
\bottomrule
\end{tabular}
\end{table*}

Table~\ref{tab:late5_appendix_neutral} shows that H-CVPO has the highest
mean reward in all 12 task--limit settings, with mean costs below the limits
in nine. On HardMove, rewards are $2.14$--$2.15$ and costs are
$0.27$--$0.60$ across all three limits. Three settings differ from the
final-policy evaluation in whether mean cost exceeds the limit. On HardMove
at $d=1.5$, the late-stage mean cost is $0.27$, below the limit, while the
final-policy mean cost is $2.17$, above it. On Sliding at $d=1.0$ and
Parking at $d=1.5$, the late-stage mean costs are $3.21$ and $2.88$,
respectively, both above the limits. Their corresponding final-policy mean
costs are $0.67$ and $0.85$, both below the limits. Parking at $d=1.0$
remains above the limit in both summaries.
The cost standard deviations of $4.35$ on Sliding at $d=1.0$ and $3.24$
on Parking at $d=1.5$ also show substantial differences across training seeds.

The broader reward--cost pattern remains similar to the final-policy results.
CHPO's mean costs are $6.13$--$10.40$ on HardMove and $152.82$ on Parking,
above every tested limit. HPPO-Lag stays below the limits on Moving,
Sliding, and Parking, but all its mean rewards on these tasks are negative;
its HardMove costs exceed $d=1.0$ and $1.5$. PDQN-Rco stays below all
three limits on Moving, but exceeds them on HardMove. Its Parking costs
at $d=1.5$ and $2.0$ are $35.62$ and $54.28$, respectively, with negative
mean rewards.

Overall, the late-stage averages support the main reward--cost comparison
beyond the final checkpoint alone. They include the final policy but also
average over four earlier checkpoints and use different evaluation episodes.
Differences between the two summaries therefore do not reflect training
changes alone. The final-policy evaluation remains the basis for assessing
policy feasibility.

\subsubsection{Extended-Training Comparison with CHPO}
\label{app:chpo_extended_budget}

We examine whether the reward--cost differences between H-CVPO and CHPO
persist when both methods are trained for more steps and CHPO's update
allocation is varied. In CHPO, the C/A ratio is the fraction of policy-update
iterations devoted to cost minimization and therefore controls the
reward--cost update balance~\citep{Zhou2025CHPO}. We extend both methods to
960k environment steps under the primary cost limit $d=1.0$ and evaluate
CHPO with C/A ratios $1/10$, $1/5$, $1/4$, and $1/3$, where $1/4$ is used
for CHPO in the main-text comparison in Figure~\ref{fig:main_comparison}.
At 320k steps, these runs recover the corresponding
trajectories from the main experiment and thus provide direct continuations
of the matched-budget comparison.

\begin{figure*}[t]
    \centering
    \includegraphics[width=\textwidth]
    {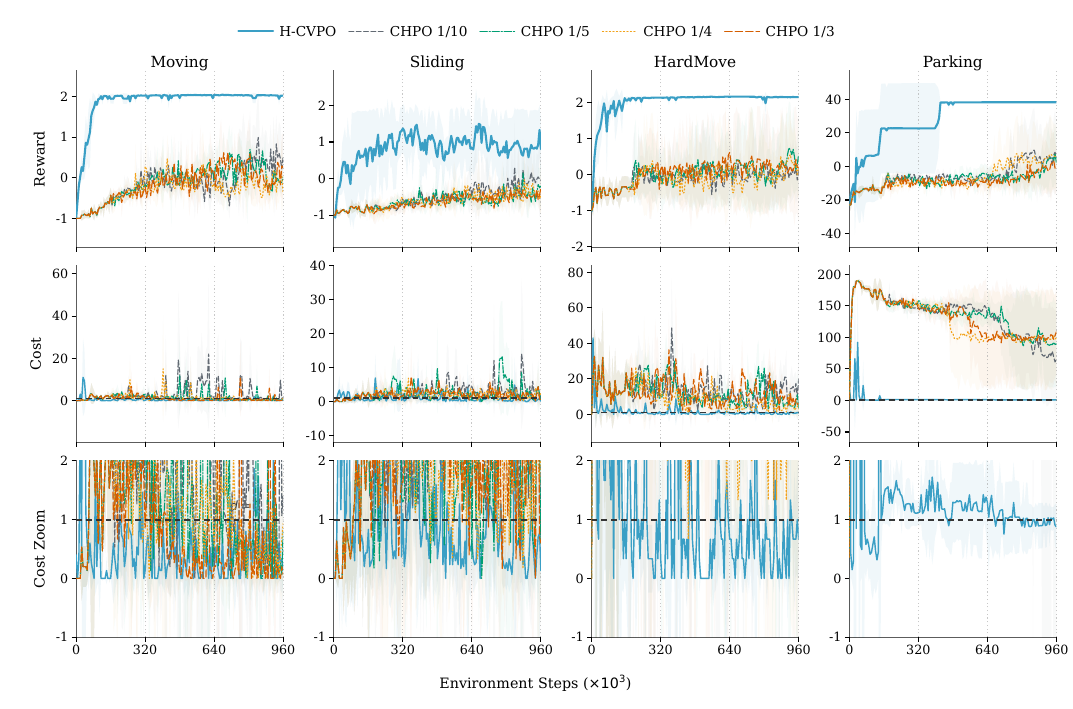}
    \caption{
    Extended-budget comparison between H-CVPO and CHPO under $d=1.0$ over
    960k environment steps. CHPO uses C/A ratios $1/10$, $1/5$, $1/4$, and
    $1/3$. Lines and shaded regions show the mean and one standard deviation
    over the three training seeds. Rows report reward, cost on the full scale,
    and cost near the constraint boundary; vertical dotted lines mark 320k,
    640k, and 960k steps, and horizontal dashed lines mark the cost limit.
    }
    \label{fig:hcvpo_chpo_960k}
\end{figure*}

Figure~\ref{fig:hcvpo_chpo_960k} shows that the separation between H-CVPO and
CHPO persists as training is extended. On Moving, H-CVPO reaches a reward
near $2.0$ within roughly the first $120$k steps and maintains this level
thereafter; following the initial transients, its mean cost fluctuates mostly
around or below the constraint boundary. HardMove follows a similar pattern,
approaching a reward of $2.0$ by approximately $160$k steps and remaining
slightly above $2.0$ through 960k, while its mean cost declines toward the
boundary over the first $200$k steps and stays comparatively low thereafter.

Parking improves in two distinct stages: its reward first rises to about $22.0$
by roughly $160$k steps, then increases to approximately $38.0$ around $420$k
and remains near that level. Its mean cost settles around $1.2$--$1.5$ over
much of the middle training period before approaching $1.0$ near the end.
Sliding exhibits a different trade-off. Its reward rises to approximately
$1.0$ by $240$k steps and reaches roughly $1.3$ around $320$k, before
fluctuating mostly between $0.6$ and $1.4$. Despite substantial variation,
its mean cost falls from about $1.1$ at 320k to near $0.1$ at 960k.

The CHPO trajectories vary substantially with the C/A ratio, and no ratio
provides a consistent reward--cost advantage across the four tasks. On Moving
and Sliding, some configurations exhibit late-stage mean costs that oscillate
around the constraint boundary, but their rewards remain below those of
H-CVPO. On
HardMove and Parking, all four configurations maintain mean costs well above
the limit through 960k steps despite gradual reward improvement, with the
largest and most persistent gap occurring on Parking.

Overall, extending training from 320k to 960k steps and varying the C/A ratio
do not eliminate the performance differences observed at 320k. The extended
trajectories instead show distinct long-horizon behavior: H-CVPO sustains high
reward as mean cost trends toward or below the constraint boundary, whereas
CHPO remains sensitive to its update allocation and retains substantial costs
on HardMove and Parking.

\subsection{Ablation Studies}
\label{app:ablation_studies}
\label{app:ablation_details}

\begin{figure}[!htbp]
    \centering
    \includegraphics[width=\textwidth]
    {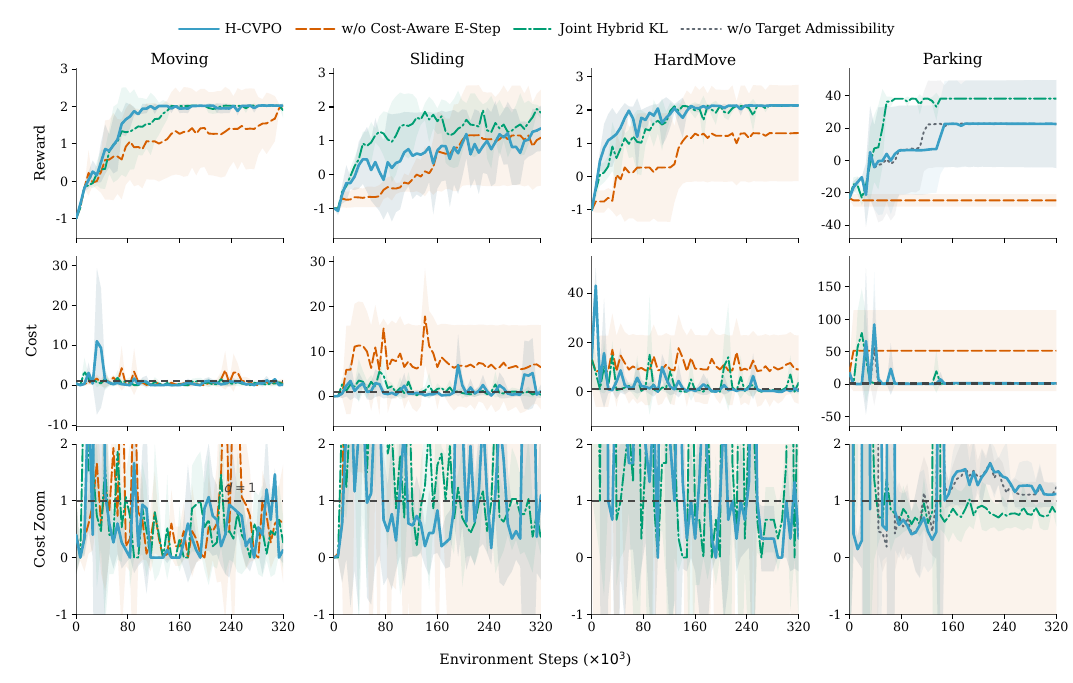}
    \caption{
    Complete ablation learning curves under the cost limit $d=1.0$.
    The variants remove the cost contribution from the E-step target,
    replace the separate discrete and conditional-continuous KL controls
    with a single joint KL constraint, or disable target-admissibility
    filtering. Curves and shaded regions show the mean and one standard
    deviation, respectively, over three training seeds; dashed lines mark
    the cost limit.
    }
    \label{fig:ablation_complete}
\end{figure}

\begin{figure}[!htbp]
    \centering
    \includegraphics[width=\textwidth]
    {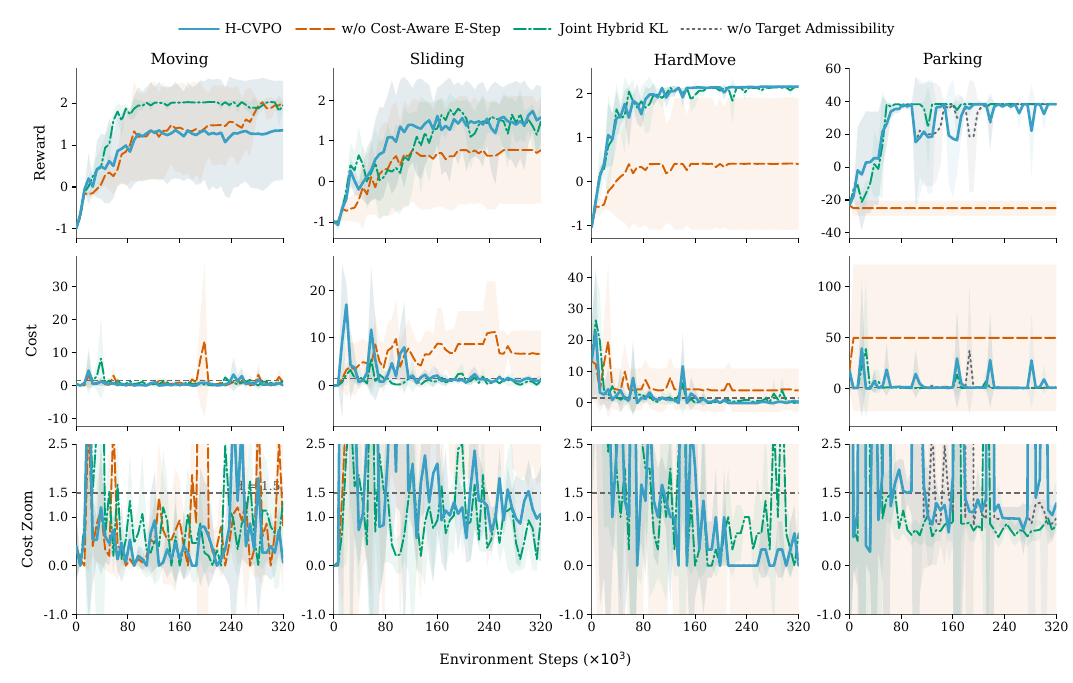}
    \caption{
    Complete ablation learning curves under the cost limit $d=1.5$.
    The variants remove the cost contribution from the E-step target,
    replace the separate discrete and conditional-continuous KL controls
    with a single joint KL constraint, or disable target-admissibility
    filtering. Curves and shaded regions show the mean and one standard
    deviation, respectively, over three training seeds; dashed lines mark
    the cost limit.
    }
    \label{fig:ablation_d1p5}
\end{figure}

\begin{figure}[!htbp]
    \centering
    \includegraphics[width=\textwidth]
    {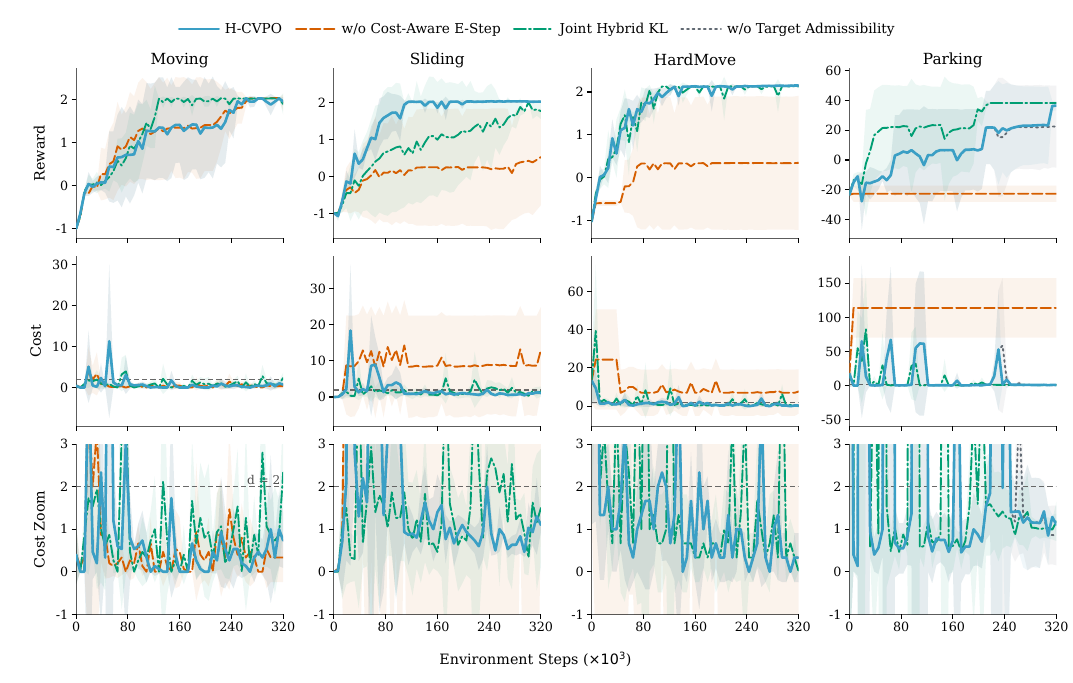}
    \caption{
    Complete ablation learning curves under the cost limit $d=2.0$.
    The variants remove the cost contribution from the E-step target,
    replace the separate discrete and conditional-continuous KL controls
    with a single joint KL constraint, or disable target-admissibility
    filtering. Curves and shaded regions show the mean and one standard
    deviation, respectively, over three training seeds; dashed lines mark
    the cost limit.
    }
    \label{fig:ablation_d2p0}
\end{figure}

We examine three components of H-CVPO: cost-aware E-step reweighting,
separate hybrid KL controls, and target-admissibility filtering.
Following Section~\ref{sec:ablation}, we compare full H-CVPO with three
variants that respectively remove the cost term from the E-step target,
replace the separate discrete and conditional-continuous M-step KL
constraints with a single joint KL constraint at the same total KL allowance,
or disable the pre-projection target-admissibility filter.
Each variant changes only one component and is evaluated under
$d\in\{1.0,1.5,2.0\}$, with all other training and evaluation settings
unchanged.
Figures~\ref{fig:ablation_complete}, \ref{fig:ablation_d1p5}, and
\ref{fig:ablation_d2p0} show the full reward and cost learning curves.
Table~\ref{tab:hcvpo_ablation_cross_d_final_policy} summarizes final-policy
performance across the three cost limits.

\begingroup
\newsavebox{\hcvpoCrossDTableBox}
\let\hcvpoOriginalTabular\tabular
\let\endhcvpoOriginalTabular\endtabular
\renewenvironment{tabular}[1]{%
  \begin{lrbox}{\hcvpoCrossDTableBox}%
  \hcvpoOriginalTabular{#1}%
}{%
  \endhcvpoOriginalTabular
  \end{lrbox}%
  \resizebox{\textwidth}{!}{\usebox{\hcvpoCrossDTableBox}}%
}
\begin{table*}[t]
\centering
\scriptsize
\setlength{\tabcolsep}{2.3pt}
\caption{Final-policy performance of H-CVPO and its three single-component
ablations across cost limits. The ablations are A1 (w/o Cost-Aware E-Step),
A2 (Joint Hybrid KL), and A3 (w/o Target Admissibility). Entries report the
mean and sample standard deviation across training seeds 11, 22, and 33, with
each final policy evaluated over 40 deterministic episodes. $\dagger$ marks a
mean episodic cost above the corresponding cost limit.}
\label{tab:hcvpo_ablation_cross_d_final_policy}
\begin{tabular}{llrrrrrrrr}
\toprule
 & & \multicolumn{2}{c}{H-CVPO} & \multicolumn{2}{c}{A1} & \multicolumn{2}{c}{A2} & \multicolumn{2}{c}{A3} \\
\cmidrule(lr){3-4}\cmidrule(lr){5-6}\cmidrule(lr){7-8}\cmidrule(lr){9-10}
Environment & $d$ & Reward & Cost & Reward & Cost & Reward & Cost & Reward & Cost \\
\midrule
Moving & 1.0 & $2.07\pm0.03$ & $0.85\pm0.97$ & $1.88\pm0.29$ & $1.73\pm0.68^{\dagger}$ & $2.01\pm0.04$ & $0.43\pm0.29$ & $2.07\pm0.03$ & $0.85\pm0.97$ \\
Moving & 1.5 & $1.41\pm1.18$ & $1.07\pm0.60$ & $2.02\pm0.13$ & $6.58\pm9.32^{\dagger}$ & $2.06\pm0.04$ & $1.33\pm0.97$ & $1.41\pm1.18$ & $1.07\pm0.60$ \\
Moving & 2.0 & $2.08\pm0.02$ & $1.00\pm0.43$ & $1.97\pm0.15$ & $0.87\pm0.93$ & $2.07\pm0.03$ & $1.78\pm0.51$ & $2.08\pm0.02$ & $1.00\pm0.43$ \\
\midrule
Sliding & 1.0 & $1.08\pm0.61$ & $0.67\pm0.47$ & $1.31\pm1.32$ & $0.52\pm0.45$ & $1.71\pm0.44$ & $0.91\pm0.12$ & $1.08\pm0.61$ & $0.67\pm0.47$ \\
Sliding & 1.5 & $1.55\pm0.85$ & $0.68\pm0.30$ & $0.66\pm1.28$ & $4.99\pm7.41^{\dagger}$ & $1.15\pm0.81$ & $0.55\pm0.42$ & $1.55\pm0.85$ & $0.68\pm0.30$ \\
Sliding & 2.0 & $1.99\pm0.10$ & $0.84\pm0.06$ & $0.56\pm1.34$ & $2.14\pm1.90^{\dagger}$ & $1.96\pm0.04$ & $0.83\pm0.13$ & $1.99\pm0.10$ & $0.84\pm0.06$ \\
\midrule
HardMove & 1.0 & $2.05\pm0.00$ & $0.42\pm0.52$ & $1.17\pm1.50$ & $3.25\pm4.99^{\dagger}$ & $2.04\pm0.01$ & $1.42\pm0.95^{\dagger}$ & $2.05\pm0.00$ & $0.42\pm0.52$ \\
HardMove & 1.5 & $2.06\pm0.01$ & $2.17\pm3.32^{\dagger}$ & $0.31\pm1.45$ & $9.17\pm8.43^{\dagger}$ & $1.97\pm0.14$ & $0.50\pm0.50$ & $2.06\pm0.01$ & $2.17\pm3.32^{\dagger}$ \\
HardMove & 2.0 & $2.05\pm0.02$ & $0.50\pm0.43$ & $0.30\pm1.51$ & $13.00\pm11.92^{\dagger}$ & $1.98\pm0.11$ & $0.75\pm0.66$ & $2.05\pm0.02$ & $0.50\pm0.43$ \\
\midrule
Parking & 1.0 & $23.26\pm26.94$ & $1.06\pm0.16^{\dagger}$ & $-18.30\pm4.46$ & $51.39\pm64.32^{\dagger}$ & $38.81\pm0.02$ & $0.78\pm0.19$ & $23.27\pm26.95$ & $1.07\pm0.15^{\dagger}$ \\
Parking & 1.5 & $38.82\pm0.01$ & $0.85\pm0.28$ & $-18.64\pm5.31$ & $51.75\pm72.03^{\dagger}$ & $38.83\pm0.00$ & $0.75\pm0.27$ & $38.82\pm0.01$ & $0.78\pm0.27$ \\
Parking & 2.0 & $35.92\pm5.03$ & $1.23\pm0.13$ & $-16.78\pm6.24$ & $108.05\pm32.27^{\dagger}$ & $38.81\pm0.02$ & $1.25\pm0.38$ & $23.31\pm26.87$ & $0.97\pm0.57$ \\
\bottomrule
\end{tabular}
\end{table*}
\endgroup

\paragraph{Cost-aware E-step reweighting.}
The variant labeled \emph{w/o Cost-Aware E-Step} in the figures removes
only the cost term $-\lambda_i^\star Q_c^i(s,m,a_c)$ from the E-step target in
Equation~(\ref{eq:hcvpo_teacher}), corresponding to the target-construction
step in Algorithm~\ref{alg:hcvpo_training}; the E-step itself is retained.
Figures~\ref{fig:ablation_complete}, \ref{fig:ablation_d1p5}, and
\ref{fig:ablation_d2p0} show task-dependent effects of removing cost-aware
reweighting. On Moving, the ablation learns more slowly at $d=1.0$ but
reaches a mean reward of about~$1.9$ at the final checkpoint. At $d=1.5$, the reward
curves remain close over much of training, with the ablation reaching a
higher mean reward near the end. At $d=2.0$, both methods reach similar
reward levels near the end. By contrast, on Sliding at $d=2.0$,
the ablation's mean reward rises to about~$0.5$ near the end, compared with
about~$2.0$ for full H-CVPO. On HardMove at $d=1.5$ and $2.0$, the
ablation's mean reward plateaus below~$0.5$, whereas full H-CVPO reaches
about~$2.1$. On both tasks, the ablation's mean cost remains above the
corresponding limit late in training for all three cost limits. Parking shows the
clearest deterioration: the ablation's mean reward remains negative, while
its late-stage mean cost stays around $50$ at $d=1.0$ and $1.5$ and above
$100$ at $d=2.0$. Full H-CVPO achieves positive rewards and much lower costs on
Parking, although cost spikes and periods above the limits remain.

Table~\ref{tab:hcvpo_ablation_cross_d_final_policy} reports the separate
final-policy evaluation: the ablation exceeds the cost limit in ten of the
twelve settings, compared with two for full H-CVPO. Eight settings change
from below-limit to above-limit mean costs: Moving at $d=1.0$ and $1.5$,
Sliding at $d=1.5$ and $2.0$, HardMove at $d=1.0$ and $2.0$, and Parking
at $d=1.5$ and $2.0$. Where full H-CVPO already exceeds the limit, the
ablation further raises mean cost from $2.17$ to $9.17$ on HardMove at
$d=1.5$ and from $1.06$ to $51.39$ on Parking at $d=1.0$.
Moving at $d=2.0$ and Sliding at $d=1.0$ remain below the limit under both
methods. The ablation's final-policy mean cost on Sliding at $d=1.0$ is
$0.52$, despite its above-limit late-stage curve. The evaluations use
different episodes; final-policy feasibility is assessed from the table.
These results support cost-aware reweighting as an important contributor
to constraint control, although removing it does not lead to a final-policy
cost-limit violation in every setting.

\paragraph{Separate hybrid KL controls.}
Figures~\ref{fig:ablation_complete}, \ref{fig:ablation_d1p5}, and
\ref{fig:ablation_d2p0} show mixed effects of replacing the separate controls
with a joint KL constraint at the same total allowance. On Moving at
$d=1.5$ and $2.0$, the joint-KL variant reaches a mean reward of about~$2.0$
earlier than full H-CVPO. At $d=1.5$, its mean reward remains higher
through the end of training. At $d=2.0$, full H-CVPO approaches the
joint-KL variant's mean reward near the end, leaving both at similar levels.
On Sliding, the variant generally has higher
rewards at $d=1.0$, whereas at $d=2.0$ its reward improves more slowly and
remains below full H-CVPO for most of training. HardMove shows broadly
similar reward curves under both methods across the three limits, with
cost spikes still present. On Parking at $d=1.0$ and $2.0$, the joint-KL
variant reaches positive rewards earlier and maintains higher rewards over
much of training. At $d=1.5$, both methods reach similar reward levels,
although full H-CVPO exhibits larger temporary declines in reward.

Table~\ref{tab:hcvpo_ablation_cross_d_final_policy} reports the separate
final-policy evaluation. Mean costs remain below the limits under both
methods in nine settings. On HardMove at $d=1.0$, the joint-KL variant
raises mean cost from $0.42$ to $1.42$, exceeding the limit. In contrast,
it lowers mean cost from $2.17$ to $0.50$ on HardMove at $d=1.5$ and from
$1.06$ to $0.78$ on Parking at $d=1.0$, bringing both below their limits.
On Parking at $d=1.0$, mean reward also rises from $23.26$ to $38.81$.
On Sliding at $d=1.5$, however, mean reward falls from $1.55$ to $1.15$,
while mean cost decreases from $0.68$ to $0.55$.
Overall, the final-policy mean cost is below the limit in eleven of the
twelve settings for the joint-KL variant and ten for full H-CVPO.
Thus, the joint budget meets the mean-cost limit in one more setting,
but this does not imply higher rewards or lower costs in every setting.
Neither design achieves both higher reward and lower cost across all
settings; their relative performance depends on the task and cost limit.

\paragraph{Target-admissibility filtering.}
H-CVPO assesses the approximate target $\widehat q_i$ using the
cost and KL residuals in Equation~(\ref{eq:approximate_residuals}) and skips
its projection when $\widehat q_i\notin\mathcal Q_{i,\mathrm{valid}}$. The
ablation removes only this pre-projection filter. The ablation also retains the post-update checks in
Algorithm~\ref{alg:hcvpo_training} for finiteness of the measured quantities
and compliance with the discrete and conditional-continuous M-step KL limits.
Across Figures~\ref{fig:ablation_complete}, \ref{fig:ablation_d1p5}, and
\ref{fig:ablation_d2p0}, the reward and cost curves for Moving, Sliding,
and HardMove overlap between the full method and this ablation; visible
differences appear only on Parking. At $d=2.0$, the Parking reward curves
follow similar trajectories, but separate near
the end: full H-CVPO rises to a higher reward level while the ablation
remains near its preceding plateau. The cost curves still show spikes,
so retaining the filter does not eliminate cost-limit violations during
training.

For the final policies, both methods have identical reported reward and
cost statistics on Moving, Sliding, and HardMove, as shown in
Table~\ref{tab:hcvpo_ablation_cross_d_final_policy}. On Parking at $d=1.0$,
removing the filter changes mean reward only from $23.26$ to $23.27$ and
mean cost from $1.06$ to $1.07$, with both costs above the limit.
At $d=1.5$, mean reward remains $38.82$, while mean cost decreases from
$0.85$ to $0.78$. The largest reward difference occurs at $d=2.0$:
removing the filter lowers mean reward from $35.92$ to $23.31$ and mean
cost from $1.23$ to $0.97$, with both costs below the limit.
These results show a reward benefit from retaining the filter on Parking
at $d=2.0$, but not a general reduction in final-policy mean cost.
Its role is to check targets before projection, not to guarantee better
performance.

\subsection{Computational and Reproducibility Considerations}
\label{app:computational_reproducibility}

\paragraph{CPU execution and resource allocation.}
All experiments were conducted on an AMD EPYC 9654 processor with 32 logical
CPUs allocated to the container (CPU IDs 128--159). We used process-level
parallelism with singleton CPU affinity: each independent training worker
was assigned one logical CPU. Within each worker, numerical-library
threading and PyTorch's internal CPU computation were each restricted to one
thread. This configuration allocates parallelism across independent training
processes while avoiding nested CPU threading within each worker. Our runtime
reproducibility checks also indicate that matching the algorithmic
configuration and source code alone may not guarantee identical training
trajectories when CPU execution settings differ. Therefore, reproducing the
reported results requires attention to both worker-to-CPU allocation and
thread configuration.

\paragraph{Training seeds and computational budget.}
We use three training seeds (11, 22, and 33) to maintain a consistent
comparison protocol with CHPO and the other baselines and to report
variability across independent runs. At 320k environment steps, H-CVPO
already achieves the highest mean reward in all 12 task--limit settings and
the highest feasibility count. Although CHPO shows modest late-stage reward
gains on Moving and HardMove, we extended both methods to 960k steps and
varied CHPO's C/A ratio to examine whether a longer training horizon could
narrow the gap. The extended results show that the performance separation
persists: H-CVPO sustains favorable reward--cost behavior, whereas CHPO
retains substantial cost violations on HardMove and Parking, with no C/A
ratio providing a consistent reward--cost advantage across tasks. We
therefore did not extend the training budget further. Three seeds provide a
practical balance between cross-run variability assessment, experimental
comparability, and computational cost.

\paragraph{Perspective on integer--continuous hybrid control.}
More broadly, we view the extension from continuous-action control to
integer--continuous hybrid control as an important step toward addressing a
wider class of constrained decision-making problems. Many engineered
systems require discrete operating decisions to be coordinated with
continuous control variables. We believe that developing optimization
principles that jointly accommodate these two types of decisions is an
important direction for constrained control and reinforcement learning.

\end{document}